\documentclass{article} %

\usepackage[T1]{fontenc}
\usepackage{iclr2027_conference,times}

\usepackage{amsmath,amsfonts,bm}

\def\eqref#1{equation~\ref{#1}}

\def\1{\bm{1}}

\DeclareMathAlphabet{\mathsfit}{\encodingdefault}{\sfdefault}{m}{sl}
\SetMathAlphabet{\mathsfit}{bold}{\encodingdefault}{\sfdefault}{bx}{n}

\usepackage{hyperref}
\DeclareRobustCommand{\suppref}[1]{\ref{#1}}
\usepackage{url}
\usepackage{booktabs}
\usepackage{array}
\usepackage{multirow}
\usepackage{pifont}
\usepackage[table]{xcolor}
\usepackage{tikz}
\usepackage{float}
\usepackage{xspace}
\usepackage{paralist}
\usepackage{enumitem}
\usepackage{placeins}
\usepackage{microtype}
\usepackage{caption}
\DeclareMathSizes{7.6}{7.6}{6}{5}
\newcommand{\papertablesize}{\fontsize{7.6}{9.1}\selectfont}
\newenvironment{papertabular}[1]{%
  \papertablesize\begin{tabular}{#1}%
}{\end{tabular}}

\definecolor{rankfirst}{HTML}{C6E6C0}
\definecolor{ranksecond}{HTML}{D8EACB}
\definecolor{rankthird}{HTML}{ECF6E5}
\newcommand{\rankone}[1]{\cellcolor{rankfirst}#1}
\newcommand{\ranktwo}[1]{\cellcolor{ranksecond}#1}
\newcommand{\rankthree}[1]{\cellcolor{rankthird}#1}

\newcommand{\methodname}{P\textsuperscript{3}-WAM\xspace}

\newcommand{\aff}[1]{{\normalfont \textsuperscript{#1}}}

\title{Perceive, Predict, and Plan: \\
a Unified World Action Model\\
for Egocentric Dexterous Manipulation}

\author{%
  Zirui Wu\aff{1} ~~~ Chen Guo\aff{2*} ~~~ Tianjian Jiang\aff{3} ~~~ Zeren Jiang\aff{4} ~~~ Zeming Li\aff{5} ~~~ \\
  \textbf{Lingchen Yang\aff{2} ~~~ Longteng Duan\aff{2} ~~~ Ying Xue\aff{2} ~~~ Handi Yin\aff{1} ~~~ } \\
  \textbf{Shenyuan Gao\aff{5} ~~~ Martin R. Oswald\aff{6} ~~~ Jun Zhang\aff{5} ~~~ Ping Tan\aff{5$\dagger$} ~~~ Jie Song\aff{1,5$\dagger$}}
  \\\\
  \textsuperscript{1} HKUST (GZ) ~~ \textsuperscript{2} Primatix ~~
  \textsuperscript{3} ETH Zurich ~~
  \textsuperscript{4} Visual Geometry Group, University of Oxford \\
  \textsuperscript{5} HKUST ~~~~
  \textsuperscript{6} University of St. Gallen ~~~~\\
  \textsuperscript{*} Project lead ~~~~ \textsuperscript{$\dagger$} Corresponding authors
  \\\\
  \url{https://p3-wam.pages.dev/}
}

\iclrfinalcopy %

\providecommand{\doi}[1]{\mbox{doi:\ \href{https://doi.org/#1}{#1}}}
\makeatletter
\let\paperoriginalurl\url
\renewcommand{\url}[1]{\begingroup
  \in@{arxiv.org/abs/}{#1}%
  \ifin@\mbox{\paperoriginalurl{#1}}\else
    \in@{openreview.net/forum?}{#1}%
    \ifin@\mbox{\paperoriginalurl{#1}}\else\paperoriginalurl{#1}\fi
  \fi
  \endgroup}
\makeatother

\begin{document}

\maketitle

\begin{figure}[!htbp]
  \centering
  \includegraphics[width=\textwidth]{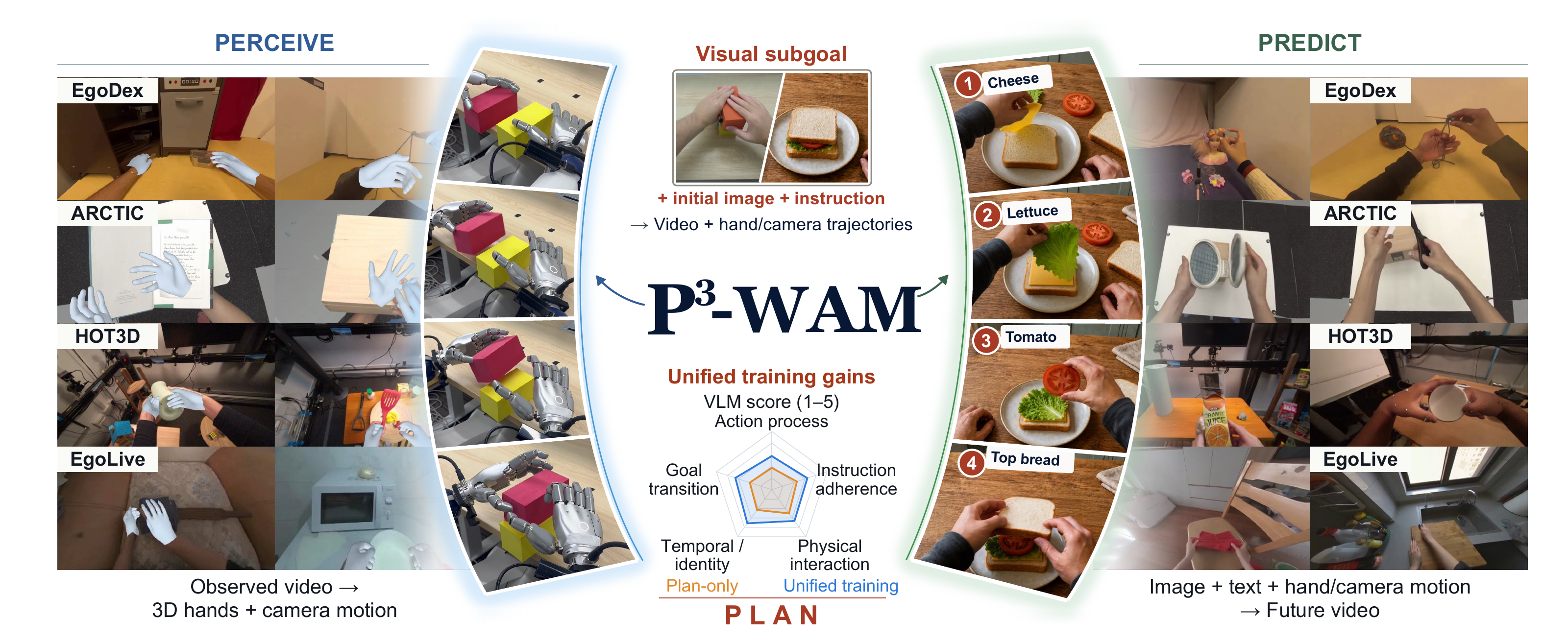}
  \caption{\textbf{\methodname unifies perception, prediction, and planning.}
  \textsc{Perceive} recovers hand motion from video;
  \textsc{Predict} generates video under prescribed motion;
  \textsc{Plan} jointly generates motion and video guided by
text instructions and optional visual goals.}
  \label{fig:teaser}
\end{figure}

\begin{abstract}
We introduce \methodname, a unified world action
model for egocentric dexterous manipulation that learns
3D perception, action-conditioned prediction, and goal-directed planning through three complementary
modes: \textsc{Perceive}, which reconstructs articulated
hand motion and camera trajectories from video;
\textsc{Predict}, which generates visual futures under
prescribed actions; and \textsc{Plan}, which jointly
generates action trajectories and their anticipated visual consequences conditioned on an initial image, a language instruction, and an optional visual goal.
Our model integrates a compact, structured 3D action representation into a pretrained video diffusion transformer through a mixture-of-transformers architecture with shared base weights, modality-specific adaptation, and bidirectional visual--action attention.
Randomized modality visibility lets video and 3D actions
serve as either conditioning context or prediction targets across the three modes.
Experiments show that unified training improves all three
capabilities over single-mode training, including gains in planning quality under a vision–language model evaluation.
Through unified training, our model outperforms specialized
baselines on 3D hand reconstruction and action-conditioned
video prediction across four datasets, while producing
goal-directed manipulation videos with stronger instruction
adherence and action progression than video-generation
baselines.
\end{abstract}

\section{Introduction}
\label{sec:introduction}

Dexterous manipulation requires more than mapping pixels directly to actions. An embodied model must answer three complementary questions: \textit{What happened in an observed interaction? What will happen under a specified action? What should be done to accomplish a desired goal?} These questions correspond to three fundamental capabilities: perceiving the 3D motion underlying observed interactions, predicting action-conditioned world dynamics, and planning actions together with their anticipated consequences. Infants can use what they learn from observing adults' actions to guide their own behavior~\citep{hauf2008actioneffect}. Through their own interactions, they learn to predict action outcomes and use those expected outcomes to select actions~\citep{verschoor2013outcome} and organize movements toward a goal~\citep{claxton2003motorplanning}. Together, these findings suggest that perception, prediction, and planning are linked by knowledge of actions and their consequences.

Despite these connections, prior work on egocentric manipulation has developed along three closely related but largely separate lines. Egocentric reconstruction methods recover structured 3D interaction information, such as hand motion and camera trajectories, from observed videos~\citep{pavlakos2024hamer, yu2025dynhamr}. Action-conditioned world models predict future observations under prescribed motion trajectories~\citep{hao2026egosim, wang2026hand2world, sun2026handworld}. World action models jointly generate actions and future observations for goal-directed control~\citep{ye2026dreamzero}. These capabilities, however, are still largely learned in isolation or coupled only in subsets.

To close this loop, we introduce \methodname, a unified world action model for egocentric dexterous manipulation that integrates three complementary modes: 3D interaction reconstruction (\textsc{Perceive}), action-conditioned world prediction (\textsc{Predict}), and joint action--future planning (\textsc{Plan}). Reconstruction develops the model's capacity for 3D perception by recovering 3D motion from observed interactions. Prediction teaches the model to anticipate the visual consequences of prescribed actions. Together, these capabilities provide the foundation for world--action planning, in which the model learns to jointly generate goal-directed hand action trajectories and their anticipated visual futures. We learn all three modes through unified training. Figure~\ref{fig:teaser} illustrates these three modes.

We build \methodname on a pretrained video diffusion transformer~\citep{wan2025video} to draw on its rich visual and motion priors. We represent each action by the articulated poses of both hands and the camera pose, and encode these states as compact tokens alongside the video tokens. We integrate this representation into the video backbone through a mixture-of-transformers (MoT) architecture with a dedicated action expert, which shares the backbone’s base weights while incorporating modality-specific low-rank adaptation (LoRA)~\citep{hu2022lora}. To train this architecture across all three modes, we randomly vary the visibility of visual observations and actions: providing video while withholding actions supports perception, providing actions while withholding future video supports prediction, and withholding actions and video supports joint action--future planning, exposing the model to complete modes and partially observed interactions. Over long sequences, successive visual subgoals specify the target arrangement for each planning window.

Unified training improves all three modes over single-mode training. Our model outperforms the evaluated baselines in \textsc{Perceive} and \textsc{Predict}. Its \textsc{Plan} videos receive higher VLM scores for action process and instruction adherence.  Robot demonstrations illustrate how generated human-hand motion can be transferred to a dexterous robot. Our main contributions are:

\begin{itemize}[nosep,leftmargin=*]
    \item A unified world action model, \methodname, for egocentric
    dexterous manipulation that integrates 3D interaction reconstruction,
    action-conditioned world prediction, and goal-guided
    world--action planning. We will release our code and model checkpoints.

    \item A unified training framework that connects pretrained video
    priors with structured 3D actions through a mixture-of-transformers
    architecture and randomized modality visibility, enabling
    complementary learning across perception, prediction, and planning.

    \item Extensive experiments indicating that perception, prediction, and planning reinforce one another through unified training, outperforming state-of-the-art baselines across all three tasks.
    Robot demonstrations further illustrate the transfer of generated human-hand motion to a dexterous robot.
\end{itemize}

\vspace{-3pt}
\section{Related Work}
\label{sec:related_work}

\vspace{-3pt}
\paragraph{Video World Models and World Action Models.}
Action-conditioned world models predict single images~\citep{sudhakar2024controllingworldsleighthand} or roll out videos~\citep{akkerman2025interdyn,dreamx2026dreamxphi,gao2026lome}. Wan~\citep{wan2025video} and Cosmos~3~\citep{nvidia2026cosmos3} provide pretrained video priors. For action representation, Dexterous World Models~\citep{kim2026dexterous} uses rendered scenes and hand-mesh trajectories, while DreamDojo~\citep{gao2026dreamdojo} learns latent actions from human videos for robot-world rollouts. Explicit scene and camera geometry can also help maintain spatial consistency. EgoSim~\citep{hao2026egosim} combines video generation with an updatable 3D scene and shared human--robot keypoints, and EgoGenesis~\citep{yan2026egogenesis} uses camera-aware 3D action embeddings and online projective memory for autoregressive rollout.

\vspace{-3pt}
World action models generate actions alongside anticipated visual dynamics through causal video prediction and inverse dynamics, as in LingBot-VA~\citep{li2026lingbotva}, or joint video--action flow matching, as in DreamZero~\citep{ye2026dreamzero}. Fast-WAM~\citep{yuan2026fastwam}, Foresight Without Seeing~\citep{huang2026forewam}, and RIFT~\citep{zhang2026rift} study predictive visual information for control without a complete iterative visual rollout at deployment.

\vspace{-3pt}
\paragraph{Monocular Hand Reconstruction.}
Recovering 3D hand motion from video requires estimating hand articulation while accounting for a moving camera. Single-image methods provide hand pose and shape estimates: HaMeR~\citep{pavlakos2024hamer} scales transformer capacity and heterogeneous training data, while WiLoR~\citep{potamias2025wilor} integrates localization, handedness estimation, and image-aligned mesh refinement. For egocentric sequences, HaWoR~\citep{zhang2025hawor} and Dyn-HaMR~\citep{yu2025dynhamr} combine hand-motion estimation with SLAM to recover global trajectories, using temporal infilling or interacting-hand priors and optimization. Concurrent work MINT~\citep{zhu2026mint} jointly estimates camera and hand motion.

\vspace{-3pt}
Video generative models offer another source of information about hand motion. Concurrent works ViDiHand~\citep{wang2026vidihand} and ACE-Ego-Hand~\citep{liu2026aceegohand} use video diffusion features for hand motion recovery. We train reconstruction jointly with future prediction and planning, allowing the model to learn from how hand interactions unfold as well as from observed video. This joint training improves 3D hand reconstruction in our single-mode comparison (Sec.~\ref{sec:ablations}).

\vspace{-3pt}
\paragraph{Unified Multimodal Models.}
A shared model can connect these tasks by changing which modalities are observed and which are generated. Unified World Models~\citep{zhu2025unified} couples video and action diffusion for forward, inverse, policy, and video inference. Language provides a further interface for world modeling and control: Motus~\citep{bi2025motus} combines language, video, and latent actions, while MotuBrain~\citep{motubrain2026} jointly models text, video, and embodiment-specific actions. Geometric modalities extend this shared formulation to spatial reasoning. Flex-$\pi$~\citep{yan2026flexpi} includes pointmap and feature streams alongside RGB and actions, and Aether~\citep{zhu2025aether} unifies 4D reconstruction, camera-conditioned prediction, and goal-image-conditioned visual planning. \methodname{} applies joint reconstruction, prediction, and planning to articulated hand interaction. It plans hand and camera trajectories together with video and supports visual goals to preserve spatial targets across a task.
\begin{figure}[!t]
\centering
  \vspace{-4pt}
\includegraphics[width=0.8\textwidth]{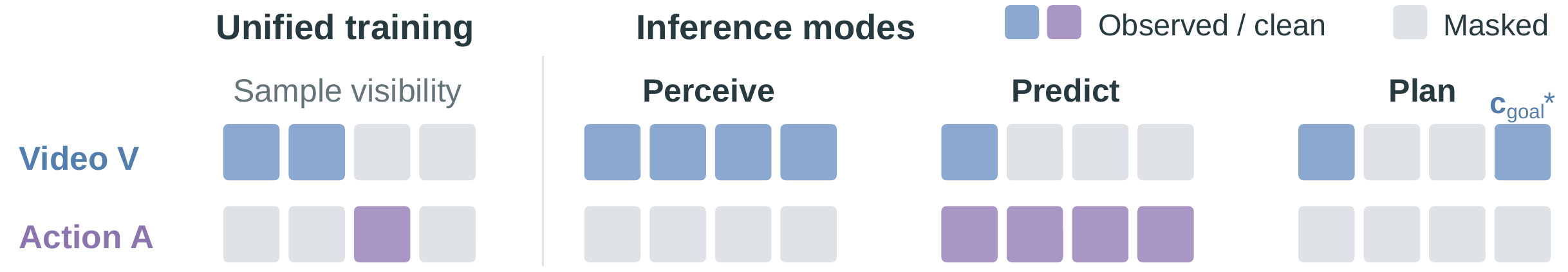}
\caption{Task modes and unified training. Colored tiles denote observed
video $\mathbf{V}$ and actions $\mathbf{A}$; gray tiles denote missing inputs.
Unified training samples these patterns across clips and time.}
\label{fig:unified-formulation}
  \vspace{-4pt}
\end{figure}

\vspace{-3pt}
\section{Method}
\label{sec:method}
\vspace{-3pt}
\subsection{Unified Task Formulation}
\label{sec:problem_formulation}

We consider an egocentric video $\mathbf{V}=\{\mathbf{I}_t\}_{t=1}^{T}$
with aligned hand states $\mathbf{H}=\{\mathbf{h}_t\}_{t=1}^{T}$ and
camera states $\mathbf{P}=\{\mathbf{P}_t\}_{t=1}^{T}$, which encode rotation and translation relative to the first frame.
We define the 3D action sequence as $\mathbf{A}=\{\mathbf{a}_t\}_{t=1}^T=\{\langle\mathbf{h}_t,\mathbf{P}_t\rangle\}_{t=1}^T$.
Action tokens $\mathbf{z}_a$ encode the aligned hand and camera observations.
The text instruction $\mathbf{c}_{\mathrm{text}}$ and initial image
$\mathbf{c}_{\mathrm{ref}}=\mathbf{I}_1$ are always given.
For planning, the goal image $\mathbf{c}_{\mathrm{goal}}=\mathbf{I}_T$
specifies the desired terminal observation. 

Figure~\ref{fig:unified-formulation} defines the three modes through
the information supplied to the same model.
\textsc{Perceive} recovers hands and camera motion from a complete video.
\textsc{Predict} generates future video given hand and camera trajectories.
\textsc{Plan} generates video together with those trajectories guided by the text instruction and the optional visual goal.
During training, we sample visibility patterns per clip and across
temporal positions. We keep observed video latents clean, add noise to frames that the model must
generate, and represent missing actions with learnable query embeddings.

\vspace{-3pt}
\subsection{Joint Video--Action Modeling}
\label{sec:unified_model}
\begin{figure}[!t]
\centering
  \vspace{-4pt}
\includegraphics[width=\textwidth]{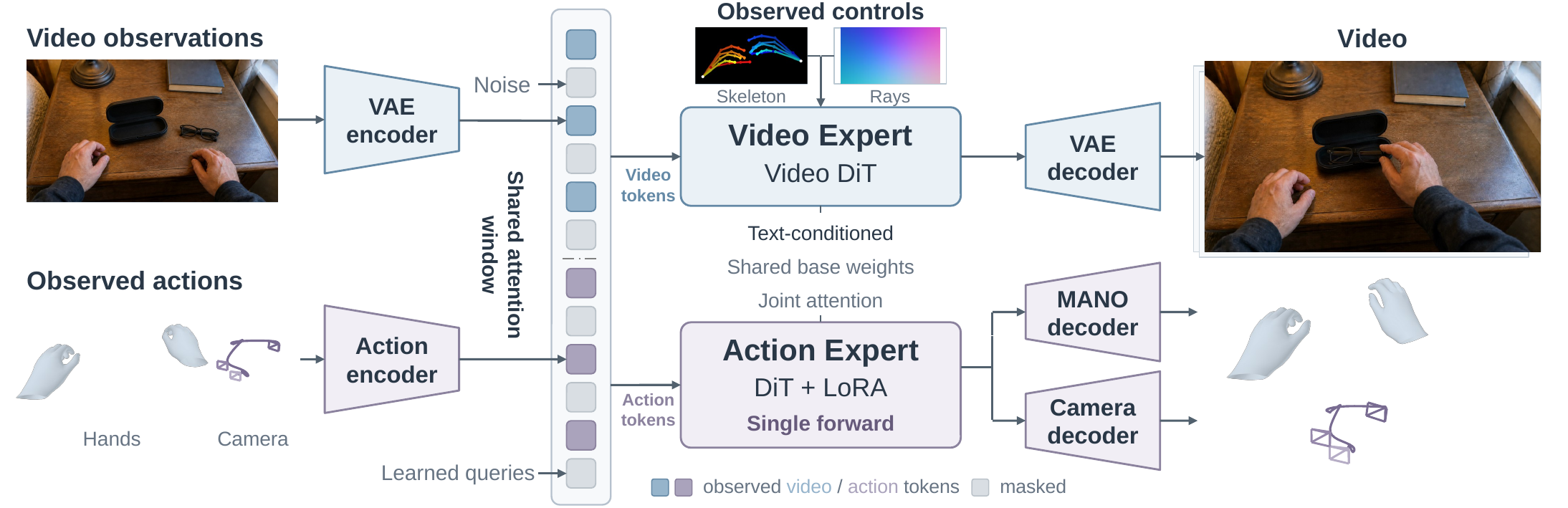}
\caption{\methodname couples video and action streams through joint attention
and shared base weights. The action expert adds low-rank adaptation and
directly predicts hand and camera states.}
\label{fig:method-overview}
  \vspace{-4pt}
\end{figure}
We build the model on top of the pretrained Wan2.2-Fun-A14B-Control model~\citep{alibabapai2025wan22fun}. We retain its latent video representation, text and first-frame conditioning, skeleton-based control pathway, and high- and low-noise diffusion experts as the RGB generation backbone.

Observed actions enter through two complementary paths: rendered hand skeletons and camera-ray maps provide spatially aligned controls for hand placement and viewpoint, while compact action tokens carry the corresponding states into joint attention.

Specifically, the RGB sequence is encoded into a spatiotemporal latent using the Wan2.1 VAE~\citep{wan2025video}. The temporally aligned hand trajectory is rendered as a 2D skeleton video and encoded by the same VAE (Fig.~\ref{fig:method-overview}). The resulting control latent is concatenated with the noisy RGB latent before DiT patch embedding. For camera conditioning, we express all camera-to-world poses in the first camera coordinate frame, making the first pose the identity, and construct per-pixel Pl\"ucker ray maps~\citep{plucker1865geometry} from the relative poses and camera intrinsics. After temporal packing to the video-latent frame rate, a lightweight convolutional adapter maps the Pl\"ucker representation into the DiT patch-feature space. Its output is then added to the joint RGB--hand patch embeddings. The same conditioning interfaces are used by both diffusion experts.

\methodname represents RGB and action as coupled latent streams
(Fig.~\ref{fig:method-overview}). A separately pretrained action autoencoder, frozen during
joint training, encodes hand and camera observations into
compact tokens that are appended to the video tokens. Within each transformer block, the RGB and action experts
share the base projection weights. The action expert adds action-specific
low-rank updates and normalization. Both streams attend to the combined
RGB and action keys and values, allowing bidirectional information
exchange. We optimize both the shared base weights and action-specific adaptations.
This modality split is used within each of Wan's high- and low-noise
experts, which operate at different diffusion noise levels.
Appendix~\suppref{app:method} specifies the shared layers and token encoding.

\vspace{-3pt}
\subsection{Action Representation}
\label{sec:geometry_codec}

We represent hands with MANO~\citep{romero2017mano}, a parametric model
that maps hand pose and shape to an articulated 3D mesh. For each hand, the model predicts global and
local rotations $\hat{\mathbf{R}}_t$, sequence-level shape
$\hat{\boldsymbol{\beta}}$, camera-space root position
$\hat{\mathbf{r}}_t$, and hand-presence probability $\hat{p}_t$.
The frozen action decoder also outputs a visibility probability
$\hat{v}_{t,j}$ for each joint. A fixed, differentiable MANO layer reconstructs camera-space
joints $\hat{\mathbf{J}}_t$ from the rotations and shape,
placing the wrist at $\hat{\mathbf{r}}_t$.

Observed hand states and camera states are encoded into
compact tokens by a pretrained, frozen action autoencoder.
A shared hand encoder embeds each hand independently, while
a separate encoder embeds the camera state. Each frame is represented by three tokens: one for the camera and one for each hand. Appendix~\suppref{app:geometry_codec} specifies the detailed architecture of the action encoder.

Unlike the video branch, which generates frames through iterative
denoising, the action branch predicts the full hand and camera
trajectory in one forward pass. We therefore represent missing
actions with learned query embeddings rather than noisy action
states. Each query corresponds to a frame and one of three action
types (camera, left hand, or right hand). Through attention to the
video and observed action tokens, the model predicts the states
at these query positions.

\vspace{-3pt}
\subsection{Optimization}
\label{sec:training}

We train video generation with flow matching~\citep{lipman2023flowmatching}, restricting the loss to unobserved video latents:
\begin{equation}
\mathcal{L}_{\mathrm{video}}
=
\mathbb{E}_{\tau,\boldsymbol{\epsilon},\mathbf{m}}
\left[
\frac{w(\tau)}{\max(1,|\mathcal{U}_v|)}
\sum_{k\in\mathcal{U}_v}
\left\|
\hat{\mathbf{v}}_{\theta,k}
-
(\boldsymbol{\epsilon}_k-\mathbf{z}_{v,k})
\right\|_2^2
\right],
\end{equation}
where $\mathcal{U}_v$ contains the unobserved video-latent
elements and $w(\tau)$ is the diffusion training weight.

The action objectives supervise MANO parameters, spatial joint
positions, temporal hand motion, hand presence, and camera states.
The total objective is
\begin{equation}
\mathcal{L} = \mathcal{L}_{\mathrm{video}}
+ \mathcal{L}_{\mathrm{param}} + \mathcal{L}_{\mathrm{spatial}}
+ \mathcal{L}_{\mathrm{temporal}} + \mathcal{L}_{\mathrm{presence}}
+ \mathcal{L}_{\mathrm{camera}}.
\end{equation}
Appendix~\suppref{app:geometry-objectives} provides the further individual
loss definitions.

\vspace{-3pt}
\subsection{Inference}
\label{sec:inference}
\begin{figure}[!t]
\centering
  \vspace{-4pt}
\includegraphics[width=\linewidth]{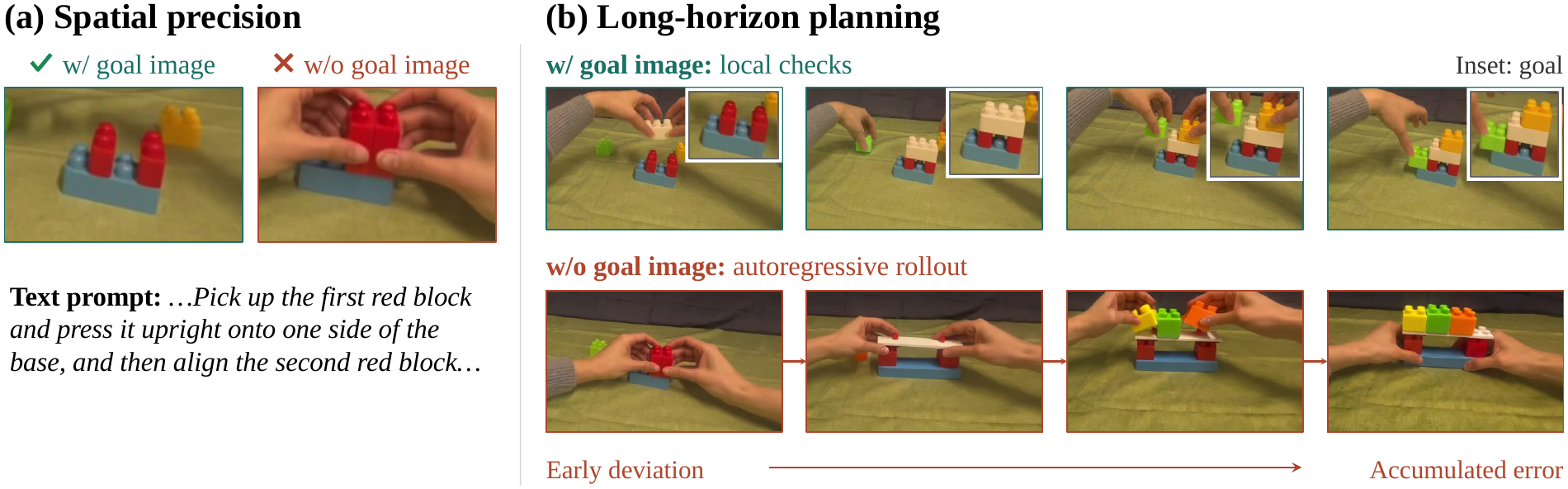}
\caption{\textbf{Visual goals specify spatial targets.}
(a) A goal image specifies how far apart the red blocks should be.
(b) Intermediate goal images guide the block arrangement at each stage
of a longer task; without them, placement errors accumulate.}
\label{fig:goal-image-control}
  \vspace{-4pt}
\end{figure}

\textsc{Perceive} reads the observed video and predicts the full action
sequence in one forward pass. In \textsc{Predict} and \textsc{Plan}, we
iteratively denoise the unobserved RGB latents while keeping observed
images fixed. For \textsc{Plan}, we keep the generated video and the action trajectory
predicted at the final denoising step.
The VAE decoder reconstructs the video frames. The MANO decoder and
fixed MANO layer recover the hand motion, and the camera decoder
recovers the camera motion.

Long tasks require several
manipulation steps, each with its own target arrangement. We use an image
generator to create visual subgoals for these steps. Each planning window
takes its starting observation, instruction, and goal image as input and
generates the video and action trajectory between the observed images.

Figure~\ref{fig:goal-image-control}(a) illustrates the spatial control
provided by a visual target: the goal image specifies the separation
between the red blocks, which the text instruction leaves ambiguous.
In the longer task in Figure~\ref{fig:goal-image-control}(b), successive
visual subgoals specify the block arrangement at each stage. The generated
sequence follows these arrangements across planning windows, while the
example without goal images accumulates placement errors and departs
from the intended structure.
\vspace{-3pt}
\section{Experiments}
\label{sec:experiments}

We first examine whether unified training benefits each mode
(Sec.~\ref{sec:ablations}), then compare \textsc{Perceive}, \textsc{Predict},
and \textsc{Plan} with specialized baselines on their respective tasks.
We also evaluate whether the planned human actions can be transferred to
a real-world dexterous robot (Sec.~\ref{sec:robot-deployment}).

\vspace{-3pt}
\subsection{Does unified training benefit each mode?}
\label{sec:ablations}
\begin{table}[t]
  \centering
  \vspace{-4pt}
  \caption{Unified versus single-mode training on \textsc{Perceive} and \textsc{Predict}.}
  \label{tab:wm-allocation-ablation}

  \setlength{\tabcolsep}{3.0pt}
  \begin{papertabular}{llrrr@{\hspace{8pt}}rrrr}
    \toprule
    & & \multicolumn{3}{c}{\textsc{Perceive} (mm)} & \multicolumn{4}{c}{\textsc{Predict} (full frame)} \\
    \cmidrule(lr){3-5}\cmidrule(lr){6-9}
    Dataset & Training setting & MPJPE $\downarrow$ & RR-MPJPE $\downarrow$ & PA-MPJPE $\downarrow$ & PSNR $\uparrow$ & SSIM $\uparrow$ & LPIPS $\downarrow$ & FVD $\downarrow$  \\
    \midrule
    EgoDex-MANO & Single mode & 52.491 & 29.170 & 12.577 & 23.106 & 0.8178 & \textbf{0.3090} & 757.06  \\
     & Unified training & \textbf{47.587} & \textbf{25.637} & \textbf{12.184} & \textbf{23.668} & \textbf{0.8182} & 0.3096 & \textbf{653.49}  \\
    \addlinespace[2pt]
    ARCTIC & Single mode & 227.608 & 27.335 & 13.144 & 12.431 & 0.5501 & 0.4940 & 1144.02  \\
     & Unified training & \textbf{150.501} & \textbf{22.946} & \textbf{12.154} & \textbf{13.431} & \textbf{0.5687} & \textbf{0.4661} & \textbf{1035.38}  \\
    \addlinespace[2pt]
    HOT3D & Single mode & 80.656 & 31.318 & 12.317 & 14.917 & 0.4151 & 0.5189 & 1466.68  \\
     & Unified training & \textbf{60.775} & \textbf{25.164} & \textbf{11.348} & \textbf{15.857} & \textbf{0.4508} & \textbf{0.5024} & \textbf{1302.88}  \\
    \addlinespace[2pt]
    EgoLive & Single mode & 93.392 & 35.905 & 13.836 & 18.968 & 0.6768 & 0.4067 & 1056.03  \\
     & Unified training & \textbf{88.728} & \textbf{32.376} & \textbf{13.238} & \textbf{19.597} & \textbf{0.6872} & \textbf{0.4008} & \textbf{1047.61}  \\
    \midrule
    Macro & Single mode & 113.537 & 30.932 & 12.969 & 17.356 & 0.6150 & 0.4321 & 1105.95  \\
     & Unified training & \textbf{86.898} & \textbf{26.531} & \textbf{12.231} & \textbf{18.138} & \textbf{0.6312} & \textbf{0.4197} & \textbf{1009.84}  \\
    \bottomrule
  \end{papertabular}
  \par\vspace{10pt}
  \begingroup
  \centering
  \caption{VLM scores for videos from \textsc{Plan}-only and unified models (1–5; higher is better).}
  \label{tab:human-vlm-evaluation}
  \setlength{\tabcolsep}{3pt}
  \begin{papertabular}{lccccc}
    \toprule
    Training setting & Action process & Instruction adherence & Physical interaction & Temporal / identity & Goal transition \\
    \midrule
    \textsc{Plan}-only & 1.88 & 2.27 & 2.52 & 2.15 & 1.98 \\
    Unified training & \textbf{2.89} & \textbf{3.22} & \textbf{3.32} & \textbf{3.57} & \textbf{3.29} \\
    \bottomrule
  \end{papertabular}
\endgroup
  \vspace{-4pt}
\end{table}
We initialize the single-mode and unified models from Wan2.2-Fun A14B-Control
and evaluate them at $5{,}000$ total training steps, with all other training
variables held fixed. Evaluation covers
EgoDex~\citep{hoque2025egodex}, ARCTIC~\citep{fan2023arctic},
HOT3D~\citep{banerjee2025hot3d}, and EgoLive~\citep{li2026egolive}.
Tables~\ref{tab:wm-allocation-ablation} and~\ref{tab:human-vlm-evaluation}
compare perception, prediction, and planning under these matched training conditions. Unified training improves both global hand placement and local articulation in \textsc{Perceive}, reducing camera-space, root-relative, and PA errors by $23.46\%$, $14.23\%$, and $5.69\%$, respectively. 
In \textsc{Predict}, PSNR increases from $17.356$
to $18.138$ dB and FVD decreases from $1105.95$ to $1009.84$, indicating gains in both pixel accuracy and video feature similarity.
The \textsc{Plan}-only baseline is trained with visual subgoals. At evaluation, both models receive the same initial image, instruction, and goal image. We use matched generation settings and score the complete video, including the goal-conditioned ending, for both models.
Unified training improves \textsc{Plan}'s VLM scores across all five
dimensions (Table~\ref{tab:human-vlm-evaluation}). The largest gains are in temporal/identity consistency ($2.15$ to $3.57$) and goal transition
($1.98$ to $3.29$), while higher action-process and physical-interaction
scores reflect more coherent manipulation sequences. Appendices~\suppref{app:training-mode-exposure}, \suppref{app:human-vlm}, and~\suppref{app:single-mode-unified} give mode exposure, evaluation details, supplementary human ratings, and qualitative comparisons.

\vspace{-3pt}
\subsection{\textsc{Perceive} Evaluation}
\label{sec:perceive-evaluation}
We compare video-based 3D hand reconstruction with HaMeR~\citep{pavlakos2024hamer}
and Dyn-HaMR~\citep{yu2025dynhamr} on the same four datasets.
Following ViDiHand~\citep{wang2026vidihand}, we report
presence-aware joint errors in camera space, after wrist alignment (RR),
and after Procrustes alignment (PA) to assess global placement and local
pose, using the protocol in Appendix~\suppref{app:perceive-metrics}.

\begin{table}[!htb]
  \centering
  \vspace{-4pt}
  \caption{Presence-aware \textsc{Perceive} reconstruction, measured by camera-space, root-relative (RR), and Procrustes-aligned (PA) MPJPE-\textit{p} in mm ($\downarrow$), and hand-visibility F1 in percent ($\uparrow$).}
  \label{tab:reconstruction-external-baselines}

  \setlength{\tabcolsep}{3.7pt}
  \begin{papertabular}{llrrrr}
    \toprule
    Dataset & Method & MPJPE-\textit{p} $\downarrow$ & RR-MPJPE-\textit{p} $\downarrow$ & PA-MPJPE-\textit{p} $\downarrow$ & F1 $\uparrow$ \\
    \midrule
    EgoDex-MANO & Ours & \textbf{32.92} & \textbf{31.58} & \textbf{13.12} & \textbf{99.96} \\
     & HaMeR & 112.92 & 44.35 & 16.22 & 91.48 \\
     & Dyn-HaMR & 115.53 & 36.66 & 18.29 & 98.17 \\
    \addlinespace[2pt]
    ARCTIC & Ours & \textbf{45.29} & \textbf{13.08} & \textbf{7.64} & \textbf{99.72} \\
     & HaMeR & 85.34 & 28.87 & 11.51 & 97.31 \\
     & Dyn-HaMR & 317.46 & 48.28 & 39.40 & 89.88 \\
    \addlinespace[2pt]
    HOT3D & Ours & \textbf{132.09} & \textbf{26.81} & \textbf{16.63} & \textbf{98.10} \\
     & HaMeR & 202.02 & 57.14 & 28.52 & 84.78 \\
     & Dyn-HaMR & 257.99 & 62.26 & 50.57 & 81.55 \\
    \addlinespace[2pt]
    EgoLive & Ours & \textbf{41.92} & \textbf{18.70} & \textbf{9.76} & \textbf{98.87} \\
     & HaMeR & 181.09 & 56.09 & 10.75 & 93.55 \\
     & Dyn-HaMR & 235.20 & 33.80 & 23.49 & 89.14 \\
    \midrule
    Macro & Ours & \textbf{63.05} & \textbf{22.54} & \textbf{11.79} & \textbf{99.16} \\
     & HaMeR & 145.34 & 46.61 & 16.75 & 91.78 \\
     & Dyn-HaMR & 231.54 & 45.25 & 32.94 & 89.68 \\
    \bottomrule
  \end{papertabular}
  \par\vspace{2pt}
  \vspace{-4pt}
\end{table}

\vspace{-3pt}
Our model achieves the lowest presence-aware reconstruction error on all
four datasets (Table~\ref{tab:reconstruction-external-baselines}). Average
camera-space MPJPE-\textit{p} is $63.05$ mm, compared with $145.34$ mm for
HaMeR and $231.54$ mm for Dyn-HaMR. Root-relative error is $22.54$ mm, compared with $46.61$ mm for
HaMeR and $45.25$ mm for Dyn-HaMR, while PA error is $11.79$ mm, compared
with $16.75$ and $32.94$ mm, respectively.
Our model also achieves the highest hand-visibility F1 on all four datasets.
Appendix~\suppref{app:perceive-unpenalized} separates joint accuracy from
missing-hand penalties, and Appendix~\suppref{app:egodex-original-joints}
evaluates reconstruction against the original EgoDex joints.

\begin{figure}[!t]
  \centering
  \vspace{-4pt}
  \includegraphics[width=0.94\textwidth]{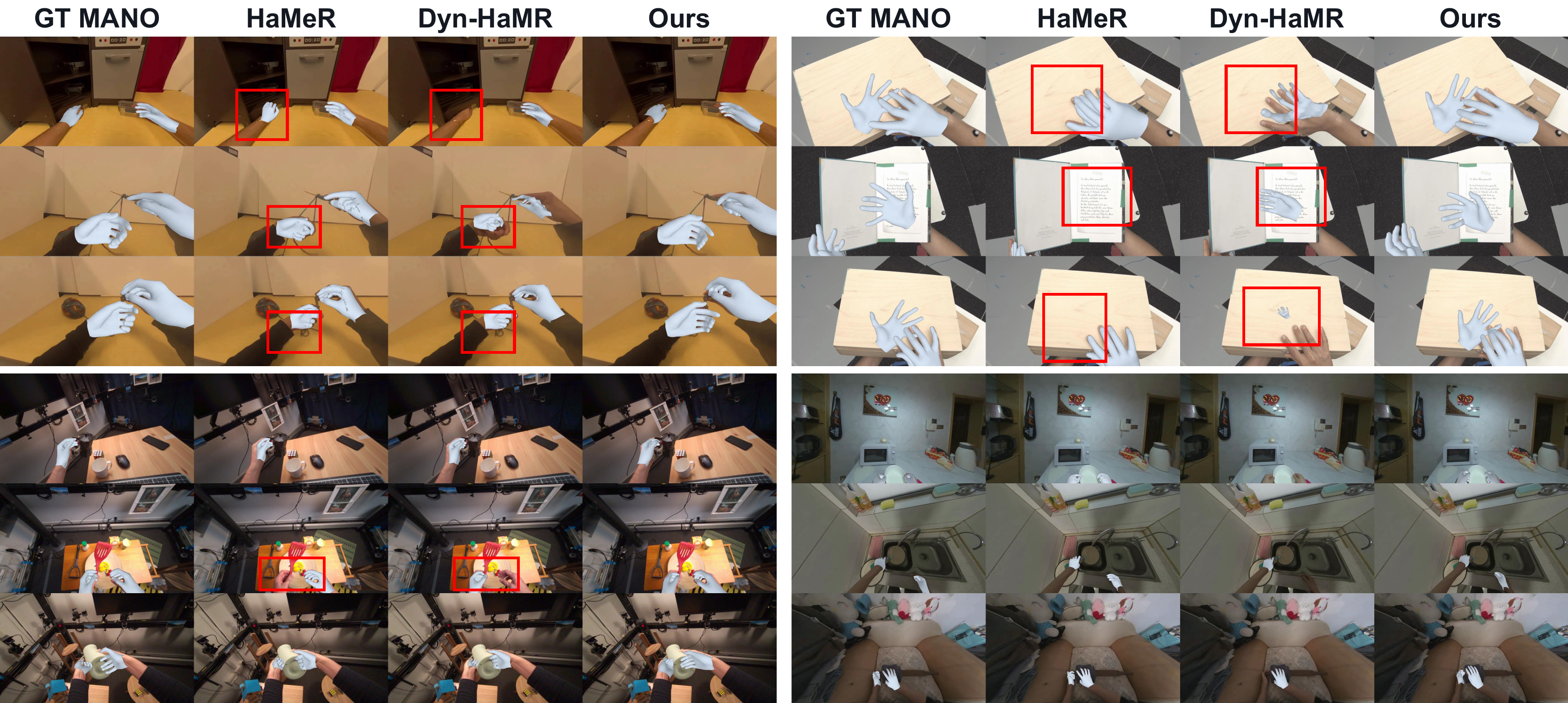}
  \caption{Qualitative \textsc{Perceive} comparisons across EgoDex, ARCTIC, HOT3D, and EgoLive. Red boxes highlight missed hands or local pose errors in baseline predictions.}
  \label{fig:recon-qualitative-comparison}
  \vspace{-4pt}
\end{figure}

\paragraph{Qualitative results.}
As can be seen in Figure~\ref{fig:recon-qualitative-comparison}, our predictions show more consistent hand
placement and articulation, while HaMeR and Dyn-HaMR miss or misplace hands, and left/right-hand confusion. We interpret our model's consistency as a benefit of combining a generative prior over plausible motions with
full-video context, since surrounding frames help resolve ambiguity when
a single frame admits several possible hand poses.

\vspace{-3pt}
\subsection{\textsc{Predict} Evaluation}
\label{sec:wm-evaluation}
\begin{figure}[!t]
  \centering
  \vspace{-4pt}
  \includegraphics[width=\textwidth]{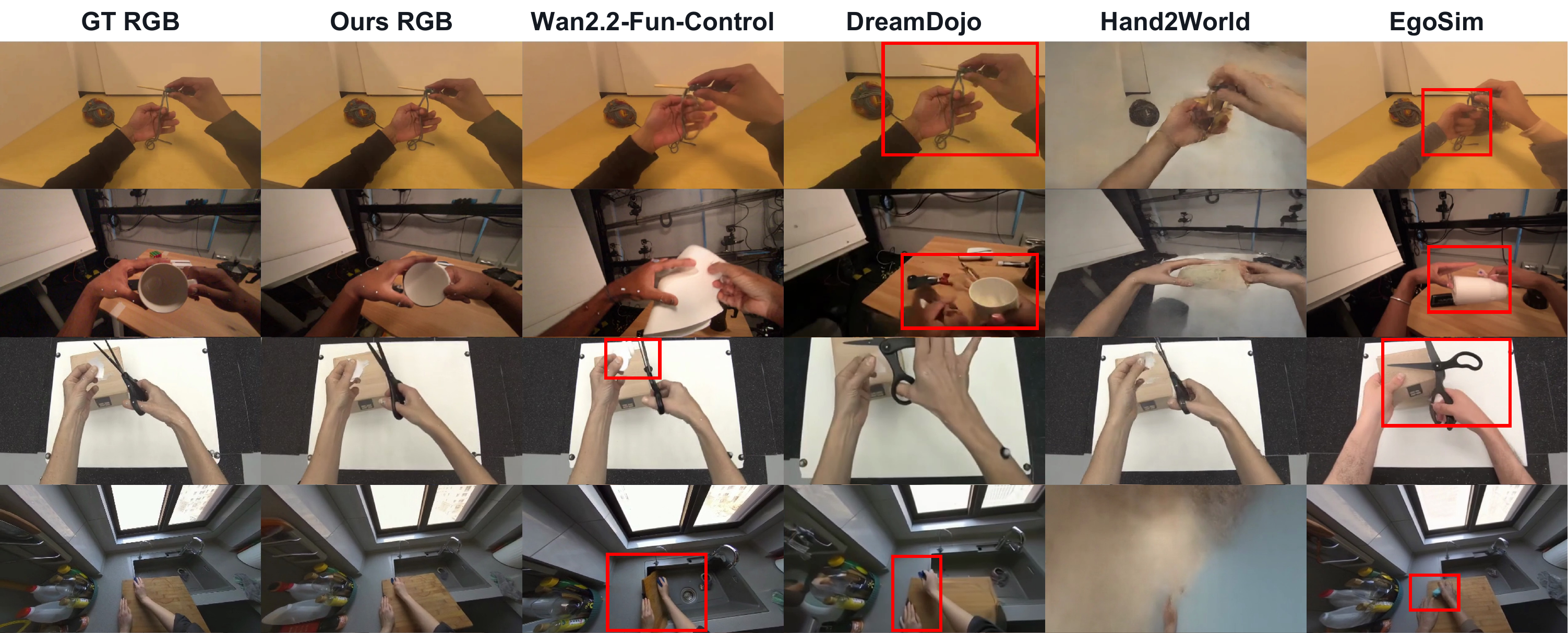}
  \caption{Qualitative \textsc{Predict} comparisons on EgoDex, HOT3D, ARCTIC, and EgoLive.}
  \label{fig:wm-qualitative-comparison}
  \vspace{-4pt}
\end{figure}
We evaluate action-conditioned video prediction on four datasets using
each method's native conditioning protocol, comparing with
Wan2.2-Fun-Control~\citep{alibabapai2025wan22fun}, DreamDojo~\citep{gao2026dreamdojo},
Hand2World~\citep{wang2026hand2world}, and EgoSim~\citep{hao2026egosim}.
Full-frame metrics measure scene fidelity, while hand-region metrics
measure the accuracy of generated hands and their placement.
Appendix~\suppref{app:predict-native-protocol} summarizes the conditioning
inputs; Appendix~\suppref{app:wm-video-metrics} details the evaluation protocol.

\begin{table}[!htb]
  \centering
  \vspace{-4pt}
  \caption{Action-conditioned video prediction across four datasets. CLIP-I measures image-feature similarity and IoU measures hand-box overlap. Darker greens mark the top three results.}
  \label{tab:world-model-external-baselines}

  \setlength{\tabcolsep}{0.55pt}
  \begin{papertabular}{@{}ll*{10}{r}@{}}
    \toprule
    & & \multicolumn{5}{c}{Full frame} & \multicolumn{5}{c}{Hand region} \\
    \cmidrule(lr){3-7}\cmidrule(lr){8-12}
    Dataset & Method & PSNR $\uparrow$ & SSIM $\uparrow$ & LPIPS $\downarrow$ & CLIP-I $\uparrow$ & FVD $\downarrow$ & PSNR $\uparrow$ & SSIM $\uparrow$ & LPIPS $\downarrow$ & CLIP-I $\uparrow$ & IoU $\uparrow$ \\
    \midrule
    EgoDex & Ours & \rankone{\textbf{24.511}} & \rankone{\textbf{0.8462}} & \rankone{\textbf{0.2679}} & \rankone{\textbf{0.9448}} & \rankone{\textbf{401.26}} & \rankone{\textbf{21.498}} & \rankone{\textbf{0.7265}} & \rankone{\textbf{0.3362}} & \rankone{\textbf{0.9203}} & \rankone{\textbf{0.7082}} \\
     & Wan2.2-Fun-Control & \rankthree{20.304} & 0.7570 & \rankthree{0.3682} & \ranktwo{0.9254} & \rankthree{719.66} & \rankthree{18.016} & \rankthree{0.6342} & \rankthree{0.4488} & \rankthree{0.8923} & \rankthree{0.6306} \\
     & DreamDojo & 19.561 & \rankthree{0.7583} & 0.3781 & 0.8868 & 931.96 & 17.498 & 0.6285 & 0.4842 & 0.8537 & 0.3005 \\
     & Hand2World & 13.287 & 0.6639 & 0.6656 & 0.7894 & 3028.61 & 12.290 & 0.5506 & 0.6580 & 0.8011 & 0.5296 \\
     & EgoSim & \ranktwo{21.331} & \ranktwo{0.8189} & \ranktwo{0.2955} & \rankthree{0.9242} & \ranktwo{649.87} & \ranktwo{19.447} & \ranktwo{0.6982} & \ranktwo{0.3647} & \ranktwo{0.9078} & \ranktwo{0.6474} \\
    \addlinespace[2pt]
    ARCTIC & Ours & \rankone{\textbf{14.106}} & \rankone{\textbf{0.5785}} & \rankone{\textbf{0.4411}} & \rankone{\textbf{0.8988}} & \rankone{\textbf{778.41}} & \rankone{\textbf{14.025}} & \rankthree{0.5951} & \rankone{\textbf{0.4602}} & \rankone{\textbf{0.8571}} & \ranktwo{0.7213} \\
     & Wan2.2-Fun-Control & \rankthree{12.875} & \rankthree{0.5560} & \ranktwo{0.4681} & \rankthree{0.8854} & \rankthree{1576.14} & 13.229 & 0.5793 & \rankthree{0.4753} & \ranktwo{0.8403} & \rankone{\textbf{0.7262}} \\
     & DreamDojo & 10.811 & \rankthree{0.5560} & 0.5677 & 0.7864 & 2430.63 & 11.246 & 0.5421 & 0.5867 & 0.7605 & 0.3128 \\
     & Hand2World & \ranktwo{12.915} & \ranktwo{0.5752} & \rankthree{0.4837} & \ranktwo{0.8873} & 2012.16 & \ranktwo{13.910} & \rankone{\textbf{0.6165}} & \ranktwo{0.4650} & 0.8188 & 0.5644 \\
     & EgoSim & 12.874 & 0.5545 & 0.4968 & 0.8840 & \ranktwo{1525.60} & \rankthree{13.780} & \ranktwo{0.5972} & 0.5084 & \rankthree{0.8301} & \rankthree{0.7035} \\
    \addlinespace[2pt]
    HOT3D & Ours & \rankone{\textbf{15.912}} & \rankone{\textbf{0.4637}} & \rankone{\textbf{0.4990}} & \rankone{\textbf{0.8845}} & \rankone{\textbf{761.35}} & \rankone{\textbf{15.390}} & \rankone{\textbf{0.4948}} & \rankone{\textbf{0.4991}} & \rankone{\textbf{0.8675}} & \rankone{\textbf{0.5921}} \\
     & Wan2.2-Fun-Control & 12.521 & 0.3299 & \ranktwo{0.6107} & \ranktwo{0.8490} & \ranktwo{1458.29} & 11.869 & \rankthree{0.4025} & \ranktwo{0.6036} & \ranktwo{0.8357} & \ranktwo{0.5095} \\
     & DreamDojo & 12.152 & \rankthree{0.3679} & 0.6509 & 0.7270 & \rankthree{1629.88} & 11.628 & 0.3632 & 0.6727 & 0.7698 & 0.1366 \\
     & Hand2World & \ranktwo{13.211} & \ranktwo{0.4065} & 0.6448 & 0.7608 & 2769.91 & \ranktwo{12.826} & \ranktwo{0.4644} & \rankthree{0.6282} & 0.7825 & \rankthree{0.4975} \\
     & EgoSim & \rankthree{12.906} & 0.3266 & \rankthree{0.6389} & \rankthree{0.8002} & 1915.92 & \rankthree{12.745} & 0.3894 & 0.6344 & \rankthree{0.7985} & 0.2590 \\
    \addlinespace[2pt]
    EgoLive & Ours & \rankone{\textbf{21.551}} & \rankone{\textbf{0.7302}} & \rankone{\textbf{0.3483}} & \rankone{\textbf{0.9373}} & \rankone{\textbf{545.89}} & \rankone{\textbf{22.316}} & \rankone{\textbf{0.6744}} & \rankone{\textbf{0.3443}} & \rankone{\textbf{0.9030}} & \rankone{\textbf{0.6621}} \\
     & Wan2.2-Fun-Control & \rankthree{18.009} & \rankthree{0.6582} & \rankthree{0.4410} & \ranktwo{0.9150} & \rankthree{1116.16} & \ranktwo{19.909} & \ranktwo{0.6265} & \ranktwo{0.4130} & \ranktwo{0.8902} & \ranktwo{0.6050} \\
     & DreamDojo & \ranktwo{18.764} & \ranktwo{0.6977} & \ranktwo{0.4359} & \rankthree{0.8747} & \ranktwo{914.66} & \rankthree{19.409} & \rankthree{0.6185} & \rankthree{0.4958} & \rankthree{0.8546} & \rankthree{0.3233} \\
     & Hand2World & 13.229 & 0.6378 & 0.5882 & 0.7238 & 3700.35 & 13.377 & 0.5742 & 0.5766 & 0.7988 & 0.1479 \\
     & EgoSim & 14.423 & 0.5459 & 0.5582 & 0.8531 & 1749.79 & 16.631 & 0.5249 & 0.5842 & 0.8311 & 0.1697 \\
    \midrule
    Macro & Ours & \rankone{\textbf{19.020}} & \rankone{\textbf{0.6547}} & \rankone{\textbf{0.3891}} & \rankone{\textbf{0.9163}} & \rankone{\textbf{621.73}} & \rankone{\textbf{18.307}} & \rankone{\textbf{0.6227}} & \rankone{\textbf{0.4099}} & \rankone{\textbf{0.8870}} & \rankone{\textbf{0.6709}} \\
     & $\Delta$ & $+3.093$ & $+0.0597$ & $-0.0829$ & $+0.0227$ & $-595.83$ & $+2.551$ & $+0.0621$ & $-0.0752$ & $+0.0223$ & $+0.0531$ \\
    \bottomrule
  \end{papertabular}
  \vspace{-4pt}
\end{table}

\vspace{-3pt}
Our model achieves the best full-frame results on all four datasets
(Table~\ref{tab:world-model-external-baselines}). Compared with our
Wan2.2-Fun-Control backbone, average PSNR rises from $15.927$ to $19.020$ dB
and FVD falls from $1217.56$ to $621.73$. Hand-region LPIPS decreases
from $0.4852$ to $0.4099$, and hand-box IoU
increases from $0.6178$ to $0.6709$. Our model leads on all five hand-region
metrics in EgoDex, HOT3D, and EgoLive, and on hand PSNR, LPIPS, and CLIP-I
in ARCTIC.

\paragraph{Qualitative results.}
At matched frames, our model preserves the manipulated objects while
following the prescribed hand--camera motion
(Figure~\ref{fig:wm-qualitative-comparison}). DreamDojo changes object shape
or identity, and EgoSim distorts hands and objects in the cup and scissors
examples. In the cutting-board example,
Wan2.2-Fun-Control deviates from the prescribed motion as the viewpoint
changes. Hand2World also shifts non-ARCTIC scenes toward ARCTIC-like
backgrounds. The comparisons suggest that explicit hand and camera
conditioning helps preserve the scene as the interaction progresses.

\vspace{-3pt}
\subsection{\textsc{Plan} Evaluation}
\label{sec:plan-evaluation}
\textsc{Plan} jointly generates a manipulation video and hand and camera
trajectories.
We compare \textsc{Plan}'s generated videos with Wan2.2 I2V-A14B~\citep{wanteam2025wan22} and
Cosmos 3-Nano~\citep{nvidia2026cosmos3}. A method-blind VLM judge
(Qwen3.8-Max~\citep{qwen2026qwen38max}) rates action process, instruction adherence,
physical interaction, temporal/identity consistency, and goal transition
on a 1--5 scale. The rubric assesses the intermediate action sequence
as well as progress toward the goal. Appendices~\suppref{app:plan-vlm-five-method}
and~\suppref{app:plan-goal-ablation} give the scoring protocol, uncertainty
estimates, and no-goal ablation.
Our generated videos receive higher VLM scores than all four baseline
settings in action process, instruction adherence, physical interaction,
and goal transition (Table~\ref{tab:plan-vlm-external-baselines}).
\begin{table}[!htb]
  \centering
  \vspace{-4pt}
  \caption{VLM comparison of \textsc{Plan}-generated videos with video-generation baselines. $\Delta$ compares Ours with the best baseline score for each metric across the four baseline settings.}
  \label{tab:plan-vlm-external-baselines}

  \setlength{\tabcolsep}{5pt}
  \begin{papertabular}{lcccccc}
    \toprule
    Method & \shortstack{Goal\\image} & \shortstack{Action\\process}
      & \shortstack{Instruction\\adherence}
      & \shortstack{Physical\\interaction}
      & \shortstack{Temporal / identity\\consistency}
      & \shortstack{Goal\\transition} \\
    \midrule
    Cosmos 3-Nano & \ding{51} & 2.279 & 2.703 & 3.190 & 2.852 & 2.080 \\
    Cosmos 3-Nano & \ding{55} & \rankthree{2.725} & 2.817 & \ranktwo{4.146} & \rankone{\textbf{4.684}} & \rankthree{2.732} \\
    Wan2.2 I2V-A14B & \ding{51} & 2.647 & \rankthree{2.847} & 3.251 & 2.741 & 2.273 \\
    Wan2.2 I2V-A14B & \ding{55} & \ranktwo{3.120} & \ranktwo{3.119} & \rankthree{4.119} & \rankthree{4.406} & \ranktwo{3.178} \\
    \midrule
    Ours & \ding{51} & \rankone{\textbf{3.591}} & \rankone{\textbf{3.739}} & \rankone{\textbf{4.350}} & \ranktwo{4.651} & \rankone{\textbf{3.841}} \\
    $\Delta$ & -- & $+0.471$ & $+0.620$ & $+0.205$ & $-0.033$ & $+0.663$ \\
    \bottomrule
  \end{papertabular}
  \vspace{-4pt}
\end{table}

\begin{figure}[!t]
  \centering
  \vspace{-4pt}
  \includegraphics[width=\textwidth]{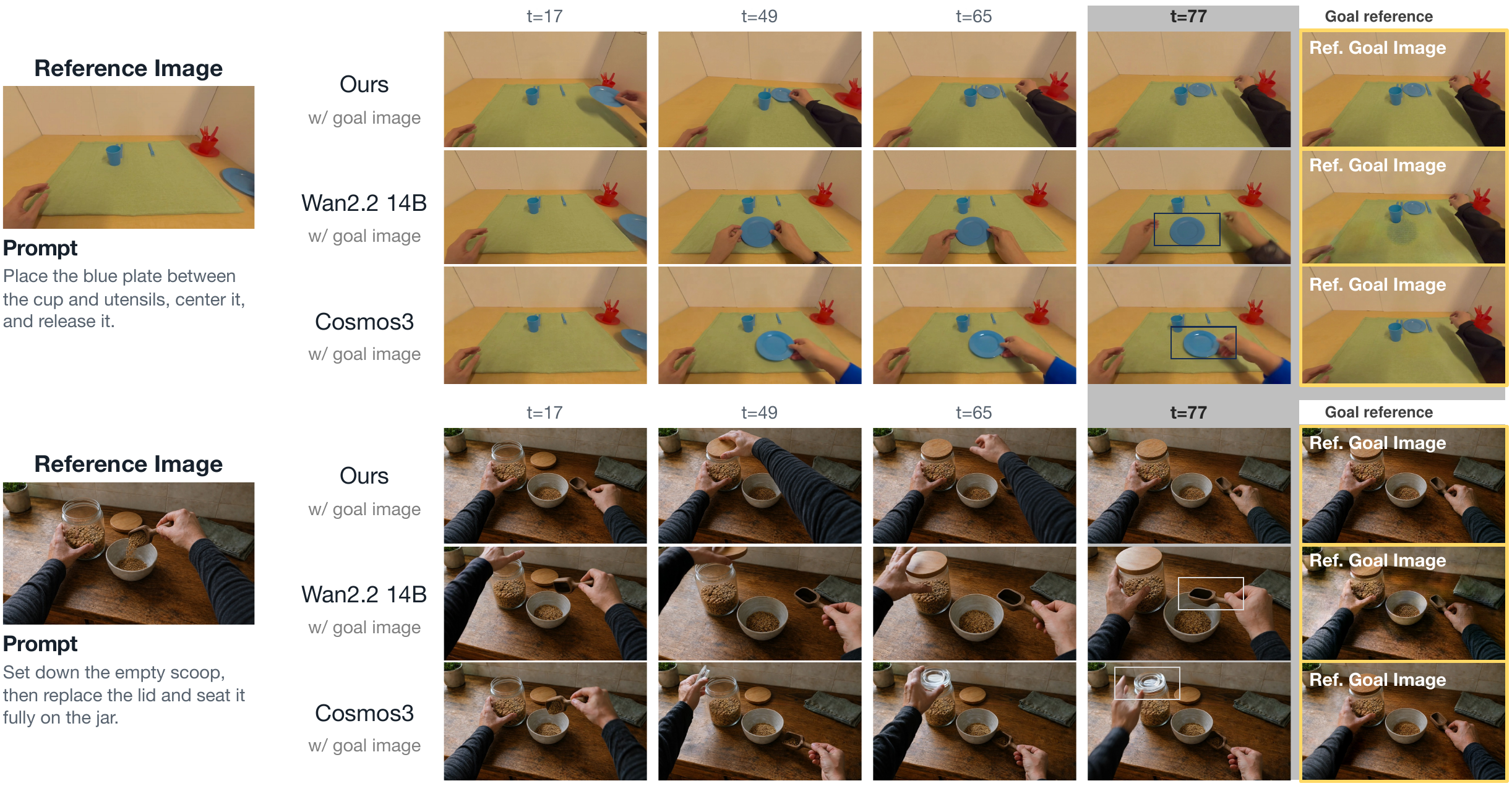}
  \caption{Goal-conditioned \textsc{Plan} comparisons.}
  \label{fig:plan-goal-conditioning-qualitative}
  \vspace{-4pt}
\end{figure}

\paragraph{Qualitative results.}
Figure~\ref{fig:plan-goal-conditioning-qualitative} compares goal-conditioned
planning. Our model places and releases the
plate at the target position within the generated part of the video.
At $t=77$, the last generated frame before the supplied goal frames
($t=78$--$81$), both baselines still leave the plate near the middle of
the mat. In the jar example, our model sets down the scoop and replaces
the lid, whereas Wan2.2 still holds the scoop and Cosmos changes the
wooden lid into a glass-like lid.

\begin{figure}[!t]
  \centering
  \vspace{-4pt}
  \includegraphics[width=\linewidth]{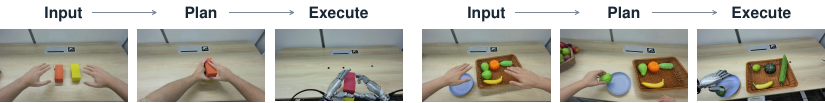}
  \caption{\textbf{Robot deployment.} Each example shows the initial observation, generated human motion, and robot execution after retargeting.}
  \label{fig:robot-deployment}
  \vspace{-4pt}
\end{figure}

\vspace{-3pt}
Appendix~\suppref{app:plan-goal-ablation} compares all methods without goal images, showing our model's hand--object consistency during manipulation. Appendix~\suppref{app:plan-geometry-consistency} measures agreement between predicted trajectories and hands reconstructed from the generated videos.
\vspace{-3pt}
\paragraph{Robot Deployment.}
\label{sec:robot-deployment}
We retarget \textsc{Plan}-generated human-hand motion to dexterous robot
commands following Do as I Do~\citep{paliwal2026doasido} and execute them
on the physical robot. Figure~\ref{fig:robot-deployment} illustrates two
examples, showing the initial observation, generated motion, and execution.

\vspace{-3pt}
\section{Conclusion}
\label{sec:conclusion}

We introduced \methodname, a unified world action model for
egocentric dexterous manipulation that integrates perception,
prediction, and planning. A coupled video--action transformer
connects visual generation with explicit hand and camera
trajectories and supports all three modes within the same model. Through extensive experiments, we show that unified training benefits hand reconstruction, action-conditioned video
prediction, and planning. 

\textbf{Limitations.}
\label{sec:limitations}
Estimating the global 3D hand position from monocular video is inherently ambiguous. Incorporating tactile or depth sensing could provide stronger geometric constraints. Our robot experiments currently rely on open-loop execution. Introducing closed-loop feedback would allow the robot to correct its motion online, potentially improving robustness over longer-horizon interactions.

\FloatBarrier
\label{sec:main-text-end}
\clearpage
\section*{AI Use Disclosure}
\label{sec:ai-use}
Generative AI assisted with manuscript drafting and editing, figure development,
and research-code development and debugging. AI-generated images also serve as
planning inputs in the examples identified as generated in the paper.
A VLM provides evaluation scores under the protocol in
Appendix~\suppref{app:human-vlm}; human ratings are collected separately.
The authors are responsible for reviewing the assisted material and for the
accuracy, originality, and attribution of the final submission.
\section*{Reproducibility Statement}

Section~\ref{sec:method} and Appendices~\suppref{app:method}--\suppref{app:implementation}
describe the architecture, objectives, codec, and training.
Appendix~\suppref{app:evaluation} specifies preprocessing, baseline interfaces,
and metrics; Appendix~\suppref{app:human-vlm} gives the human/VLM rubric and
bootstrap procedure. Additional measurements appear in
Appendix~\suppref{app:additional_quantitative}.

\newpage
\appendix
\section{Additional Method Details}
\label{app:method}
\subsection{Shared weights and modality-specific adaptation}
Within each diffusion expert, RGB and action use the same base
self-attention Q/K/V and output projections and the same two FFN
projections. The action branch adds low-rank updates to these six projections
and to the text cross-attention Q and output projections. Text keys
and values are shared. Each update has the form
$\Delta W_a=(\alpha/r)B_aA_a$, with rank $r=128$ and $\alpha=128$.
$A_a$ uses Kaiming initialization and $B_a$ starts at zero.
The pretrained base weights remain trainable.

The action branch has separate affine Q/K normalization in self-attention,
Q normalization and pre-normalization in text cross-attention, and
adaptive normalization parameters, initialized from the RGB branch.
The parameter-free layer normalizations retain the same operation in
both branches. The high- and low-noise diffusion experts each have
their own base weights and action adaptations. The video VAE, text
encoder, and action codec are frozen.

\subsection{Attention and token encoding}
Self-attention uses the concatenated RGB and action keys and values:
\[
\operatorname{Attn}(Q_m,K,V)
=\operatorname{softmax}\!\left(\frac{Q_m[K_v;K_a]^\top}{\sqrt{d}}\right)
[V_v;V_a],\qquad m\in\{v,a\}.
\]
Thus, each query normalizes over a common set of tokens from both
modalities. Learned role, handedness, and frame embeddings distinguish
the camera, left-hand, and right-hand tokens. Action RoPE uses the
fractional video-latent time $(t-1)/4$ and the center of the spatial
patch grid, with the same temporal offset as any prepended RGB reference.
Unobserved action is replaced by learned query tokens while retaining
these positions and identities. Observed RGB tokens remain clean and
receive a zero-noise timestep; the hand and camera masks also gate
their visual control inputs.

\subsection{Masked input construction}
\label{app:masked-inputs}
Mask value one denotes an observed input and zero denotes a withheld input.
The base patterns $(m_v,m_h,m_P)$ are $(1,0,0)$ for \textsc{Perceive},
$(0,1,1)$ for \textsc{Predict}, and $(0,0,0)$ for \textsc{Plan}.
The initial image and optional \textsc{Plan} goal are applied separately.

\paragraph{Temporal mode sampling.}
We use the video, hand, and camera visibility masks in
Fig.~\ref{fig:unified-formulation}, written at latent frame $\ell$ as
$(m_{v,\ell},m_{h,\ell},m_{P,\ell})$.
For the unified-training ablation, the probabilities of
\textsc{Perceive}, \textsc{Predict}, and \textsc{Plan} are
$0.2$, $0.5$, and $0.3$.
With probability $0.5$, we draw one mode for the whole clip;
otherwise, we draw modes independently at each latent position.
This exposes the model to complete tasks and temporally mixed observations.
The hand and camera masks are expanded to their temporally aligned frames.

\paragraph{Masked input construction.}
For RGB, we define the effective observation mask
$\bar{m}_{v,\ell}$ by preserving the sampled mask and
always setting $\bar{m}_{v,1}=1$.
When $\mathbf{c}_{\mathrm{goal}}\ne\emptyset$, its aligned
terminal RGB latent is also marked as observed.
Given a noise level $\tau$ and Gaussian noise
$\boldsymbol{\epsilon}_{\ell}\sim\mathcal{N}(0,\mathbf{I})$,
the video input and its per-frame noise level are
\begin{equation}
\begin{aligned}
\tilde{\mathbf{z}}_{v,\ell}
&=
\bar{m}_{v,\ell}\mathbf{z}_{v,\ell}
+
(1-\bar{m}_{v,\ell})
\big[(1-\tau)\mathbf{z}_{v,\ell}
+\tau\boldsymbol{\epsilon}_{\ell}\big],\\
\tau_{v,\ell}
&=(1-\bar{m}_{v,\ell})\tau .
\end{aligned}
\end{equation}
Thus, observed video latents remain clean and receive
clean-state timestep embeddings, while unobserved positions
receive the sampled corruption level.

For action, let $\mathbf{z}_{a,t}$ contain the encoded
camera, left-hand, and right-hand tokens at frame $t$,
and let $\mathbf{m}_{a,t}$ contain their aligned visibility
masks, broadcast over token channels.
We construct
\begin{equation}
\tilde{\mathbf{z}}_{a,t}
=
\mathbf{m}_{a,t}\odot\mathbf{z}_{a,t}
+
(1-\mathbf{m}_{a,t})
\odot\mathbf{e}^{\mathrm{query}}_{a,t},
\end{equation}
where $\mathbf{e}^{\mathrm{query}}_{a,t}$ contains learned
query tokens for unavailable action.
The same visibility masks gate the rendered-hand and
camera-control pathways.
Video flow matching is applied only to unobserved RGB
positions, while hand and camera predictions are supervised
using the available target annotations across all three modes.

\subsection{Action objectives}
\label{app:geometry-objectives}
The hand objectives supervise the MANO predictions and reconstructed joints
defined in Sec.~\ref{sec:geometry_codec}. Here, $\rho$ is the element-wise
smooth-$L_1$ penalty, $d_R$ is rotation geodesic distance, and
$\langle\cdot\rangle$ averages over valid annotations. Batch and hand
indices are omitted for clarity.

\paragraph{MANO parameters.}
We jointly supervise global orientation, local articulation,
and hand shape:
\begin{equation}
\mathcal{L}_{\mathrm{param}}
=
\left\langle
\frac{1}{N_j}\sum_{j=1}^{N_j}
\lambda_j\,d_R(\hat{\mathbf{R}}_{t,j},\mathbf{R}_{t,j})
\right\rangle
+
\lambda_{\mathrm{shape}}
\left\langle
\rho(\hat{\boldsymbol{\beta}}-\boldsymbol{\beta})
\right\rangle .
\end{equation}
Here, $N_j$ counts the global and local MANO rotations,
and $\lambda_j$ specifies their fixed supervision weights.
Shape parameters are shared across frames.

\paragraph{Spatial joint positions.}
We supervise MANO-derived joints in 3D camera space and
the image~plane:
\begin{equation}
\mathcal{L}_{\mathrm{spatial}}
=
\lambda_{\mathrm{3D}}
\left\langle
\rho(\hat{\mathbf{J}}_t-\mathbf{J}_t)
\right\rangle
+
\lambda_{\mathrm{2D}}
\left\langle
\rho\!\left(
\pi_{K_t}(\hat{\mathbf{J}}_t)-\pi_{K_t}(\mathbf{J}_t)
\right)
\right\rangle .
\end{equation}
Reference joints are reconstructed from the annotated MANO
parameters using the same layer.
The projection $\pi_{K_t}$ uses the camera intrinsic matrix
$K_t$ and normalizes coordinates by image size.

\paragraph{Temporal motion.}
We supervise articulated motion and root trajectories
at several \mbox{temporal scales:}
\begin{equation}
\mathcal{L}_{\mathrm{temporal}}
=
\lambda_{\mathrm{art}}\mathcal{L}_{\mathrm{art}}
+
\lambda_{\mathrm{traj}}\mathcal{L}_{\mathrm{traj}} .
\end{equation}
$\mathcal{L}_{\mathrm{art}}$ combines losses on local rotation
changes and root-relative joint displacements, while
$\mathcal{L}_{\mathrm{traj}}$ combines losses on root
displacements and accelerations. Both match predicted motion
to annotated motion over valid temporal intervals.

\paragraph{Hand presence.}
We supervise hand presence using binary cross-entropy:
\begin{equation}
\mathcal{L}_{\mathrm{presence}}
=
-\lambda_{\mathrm{presence}}\left\langle
p_t\log\hat{p}_t+(1-p_t)\log(1-\hat{p}_t)
\right\rangle ,
\end{equation}
averaged over known hand-level presence labels.

\paragraph{Camera parameters.}
For camera states $\mathbf{P}_t$, we supervise rotation $\mathbf{R}_t^{\mathrm{cam}}$, translation
$\mathbf{t}_t^{\mathrm{cam}}$, and field of view $\boldsymbol{\phi}_t$:
\begin{equation}
\begin{aligned}
\mathcal{L}_{\mathrm{camera}}
={}&
\lambda_{\mathrm{cam},R}
\left\langle
\|\hat{\mathbf{R}}_t^{\mathrm{cam}}-\mathbf{R}_t^{\mathrm{cam}}\|_F^2
\right\rangle
+
\lambda_{\mathrm{cam},t}
\left\langle
\|\hat{\mathbf{t}}_t^{\mathrm{cam}}-\mathbf{t}_t^{\mathrm{cam}}\|_2^2
\right\rangle + \lambda_{\mathrm{fov}}
\left\langle
\|\hat{\boldsymbol{\phi}}_t-\boldsymbol{\phi}_t\|_2^2
\right\rangle.
\end{aligned}
\end{equation}
Camera poses are expressed relative to the first frame.
The vector $\boldsymbol{\phi}_t$ contains horizontal and
vertical fields of view.

\section{Action Codec Details}
\label{app:geometry_codec}
\paragraph{Architecture and training.}
The deterministic action autoencoder represents each frame with three
64-dimensional tokens: camera, left hand, and right hand. A shared hand
encoder takes the 21-joint observation for each hand, and a separate camera
encoder takes rotation, translation, and field of view. Encoders and decoders
use residual MLPs of width 512 with three residual blocks. Training minimizes
masked reconstruction losses for joints, image coordinates, depth, visibility,
and camera states. We use AdamW with learning rate $10^{-3}$ and a global
batch of 128. Codec training takes approximately 110 H100 GPU-hours.
The codec is frozen during video--action training.

Table~\ref{tab:geometry-codec-quality} reports held-out reconstruction on
EgoDex, ARCTIC, HOT3D, and EgoLive, matching the main reporting datasets.
The multi-domain codec achieves camera-space joint errors of
$0.868$--$2.600$ mm and image-space errors of $0.323$--$0.882$ pixels.
These measurements characterize compression of annotated actions before
video-model~prediction.

\begin{table}[!htbp]
  \centering
  \caption{Multi-domain action-codec reconstruction quality on held-out validation data.
  All values are dataset means; lower is better.}
  \label{tab:geometry-codec-quality}

  \setlength{\tabcolsep}{3.5pt}
  \begin{papertabular}{lrrrrr}
    \toprule
    Evaluation & Cam. (mm) & World (mm) & RR (mm) & UV (px) & Depth (mm) \\
    \midrule
    EgoDex & 0.868 & 1.048 & 0.967 & 0.323 & 0.432 \\
    ARCTIC & 2.600 & 4.212 & 2.543 & 0.882 & 1.137 \\
    HOT3D & 1.336 & 2.088 & 1.406 & 0.580 & 0.691 \\
    EgoLive & 1.498 & 1.627 & 1.573 & 0.470 & 0.891 \\
    \bottomrule
  \end{papertabular}
\end{table}

\section{Implementation and Training Details}
\label{app:implementation}
\label{sec:implementation_details}
\paragraph{Optimization.}
We initialize the video backbone from Wan2.2-Fun-A14B-Control and train the
base transformer weights, camera adapter, action adaptations, and MANO head.
The video VAE, text encoder, and pretrained action codec remain frozen.
Training uses bfloat16, fully sharded data parallelism, and gradient
checkpointing. The final unified training recipe uses a global batch of 16. Clips contain 81 frames at 16 fps and
$848\times480$ resolution.

We use AdamW with $(\beta_1,\beta_2)=(0.9,0.999)$, weight decay $0.01$,
and a backbone learning rate of $10^{-5}$, held constant after 200 warmup
updates. We clip video gradients at norm 1 and use an exponential moving
average with decay $0.999$ for evaluation.
We train the model for approximately 7,200 GPU hours in total.

\paragraph{Training-mode exposure.}
\label{app:training-mode-exposure}
The single-mode and unified ablations use the same architecture,
initialization, data mixture, and total number of optimizer updates.
Single-mode training assigns every clip to its designated mode.
Unified training uses the mixture in Appendix~\suppref{app:masked-inputs}:
$20\%$, $50\%$, and $30\%$ of latent positions are assigned to
\textsc{Perceive}, \textsc{Predict}, and \textsc{Plan} in expectation.
The whole-clip branch accounts for $10\%$, $25\%$, and $15\%$
of clips, respectively; the other $50\%$ use independently sampled modes
across time. These are sampling probabilities, so matching total updates
does not match the exposure to any one complete mode.
Action supervision uses available annotations in every mode, while video
flow matching supervises unobserved RGB positions.
The shared transformer weights receive gradients from both objectives,
connecting action reconstruction and video generation across the sampled tasks.

\paragraph{Sampling.}
\textsc{Perceive} uses one action prediction from the clean video at zero
noise. \textsc{Predict} and \textsc{Plan} use 50 denoising steps, with the
high-/low-noise expert boundary at $\tau=0.9$. Text guidance is 6 for the
main \textsc{Predict} comparison and 2 for \textsc{Plan} evaluation.
Observed initial and goal latents stay fixed throughout sampling;
Appendix~\suppref{app:plan-goal-protocol} specifies their encoding. Paired
comparisons use matched inputs and seeds.

\section{Dataset and Evaluation Protocols}
\label{app:evaluation}
\paragraph{Data and preprocessing.}
We train on the training splits of EgoDex, EgoLive, and ARCTIC.
We apply field-of-view augmentation during training.

\paragraph{Evaluation protocol.}
Methods receive temporally aligned source observations through their native
interfaces.
Dataset means are macro-averaged with equal weights. The following sections
specify detection, joint association, and image metrics;
Appendix~\suppref{app:human-vlm} gives the \textsc{Plan} protocol and statistical
analysis.

\subsection{\textsc{Predict} conditioning}
\label{app:predict-native-protocol}
The methods use the following motion inputs.
\begin{center}
\begin{papertabular}{@{}p{.22\linewidth}p{.73\linewidth}@{}}
\toprule
Method & Motion conditioning \\
\midrule
Ours & Hand-skeleton latents, Pl\"ucker camera rays, and hand/camera tokens. \\
Wan2.2-Fun-Control & A hand-control video through its native control encoder. \\
DreamDojo & Latent actions encoded from the source video with its released action model. \\
Hand2World & MANO renders and camera parameters through its native interface. \\
EgoSim & Rendered hand skeletons and scene/mask videos along its camera path. \\
\bottomrule
\end{papertabular}
\end{center}

\subsection{\textsc{Predict} RGB and hand-region metrics}
\label{app:wm-video-metrics}
For full-frame and hand-region CLIP-I~\citep{radford2021clip}, we compute paired
image-feature cosine similarity using OpenAI ViT-B/32 with its standard
bicubic short-edge resize to 224, center crop, and normalization. FVD~\citep{unterthiner2018fvd}
uses the StyleGAN-V~\citep{skorokhodov2022styleganv} Kinetics-400 I3D TorchScript model's 400-dimensional
pre-softmax features, bilinear resizing to $224\times224$, pixel scaling to
$[-1,1]$, and one feature vector per video. We use unbiased empirical
covariances and an exact low-rank Fr\'echet-distance calculation.

Hand-region metrics use WiLoR detections with confidence threshold $0.3$
and NMS IoU threshold $0.7$. We expand ground-truth boxes by $1.5\times$,
clip them to the image boundary, and apply the same crop to the reference
and prediction. Crop PSNR, SSIM~\citep{wang2004ssim}, and VGG-LPIPS~\citep{zhang2018lpips} are measured at the native
crop resolution, with small boundary crops padded for LPIPS.

For IoU, we match unexpanded boxes by maximum total IoU within handedness
classes. We divide the matched IoU sum by the larger detection count, so
unmatched detections contribute zero. Frames with no detections on either
side are omitted; one-sided detections receive zero. Crop metrics average
hands within a frame, valid frames within a window, and then windows within
a dataset/method. Crop metrics use frames with a valid GT hand detection. FVD is computed separately for each dataset/method.
The Macro block reports the arithmetic mean of the four dataset-level
scores for each method and metric, using unrounded values and equal dataset
weights.

\subsection{\textsc{Perceive} targets and metrics}
\label{app:perceive-metrics}
Following ViDiHand~\citep{wang2026vidihand}, we report presence-aware
MPJPE-\textit{p} in camera-space, root-relative, and Procrustes-aligned
coordinates. A reference hand is eligible when at least one annotated
joint lies inside the image at camera depth greater than $0.01$ m.
This is also the positive GT definition for hand visibility.
Joint errors use the valid annotated joints of each eligible hand;
RR requires a valid wrist, and PA additionally requires at least three
joints and a \mbox{nondegenerate prediction}.

\paragraph{Association and missing-hand penalties.}
Our model emits a hand prediction when its maximum joint-visibility
probability is at least $0.5$; HaMeR and Dyn-HaMR use their released
hand detections. All methods are matched to reference hands of the same
side by descending mesh-box IoU, using a $1.1\times$ enlarged reference
box and an IoU threshold of $0.1$.
An eligible hand without a match receives a canonical MANO hand in
the penalized metrics. Its camera-space and RR penalties are the
corresponding canonical-hand errors; its PA penalty uses the canonical
RR error without Procrustes fitting. The same matching and missing-hand
penalties apply to all three methods.
Ordinary errors for each method use its own successfully matched
eligible hands, rather than a common intersection across methods.

For EgoDex-MANO, we fit MANO to the annotated 3D joints of each hand,
optimizing global orientation, finger articulation, and shape with pose
and shape regularization. The annotated wrist anchors the hand position.
We reconstruct the fitted MANO21 joints in camera coordinates and use them
as the common evaluation targets for all methods. ARCTIC, HOT3D, and
EgoLive use their provided 3D hand annotations. We average windows within
each dataset and then average the four datasets equally.

\paragraph{HaMeR camera recovery.}
For HaMeR, we recover camera-space translation using known dataset intrinsics.
Native predictions follow the official demo's crop-to-full
conversion with image-scaled focal length $5000\max(W,H)/256$
($16562.5$ pixels for our $848\times480$ inputs).
For each detected hand in each frame, let $\mathbf{q}_j$ be its predicted wrist-relative
3D joints and $\mathbf{u}_j$ its native full-image 2D projections. We fit
$\mathbf{t}$ by minimizing
$\sum_j\|\pi_K(\mathbf{q}_j+\mathbf{t})-\mathbf{u}_j\|_2^2$
using the dataset intrinsic matrix $K$, including both focal lengths and
the principal point, transformed to the evaluation image.
Rotation, articulation, shape, and metric scale remain fixed; fitting uses
only HaMeR's own 2D predictions, with reference joints used only for scoring.
We retain the native 2D detections, mesh-overlap assignments, and missed-hand
penalties, so RR, PA, and visibility scores are unchanged.
EgoLive's provided 2D and 3D annotations retain a median $1.62$-pixel
reprojection residual under the released intrinsics; we use those intrinsics
without fitting them to the reference annotations.

\paragraph{Hand visibility.}
We evaluate whether each hand is in view at each frame, using the
reference-hand eligibility defined above. For Ours, the frozen action
decoder's joint-visibility probabilities $\hat{v}_{t,j}$ give a positive
hand prediction when $\max_j\hat{v}_{t,j}\geq0.5$.
This is separate from the MANO head's hand-level presence output
$\hat{p}_t$ and its presence loss.
Baselines use their same-side hand detections whose
projected mesh boxes intersect the image. We pool these hand-frame
decisions to compute precision, recall, and F1, without an IoU matching
requirement. For Ours, these visibility decisions also select the
action predictions before mesh-overlap matching. Visibility F1
measures hand-in-frame classification; joint errors additionally
evaluate the spatially matched hand states.

\subsection{Goal-image conditioning}
\label{app:plan-goal-protocol}
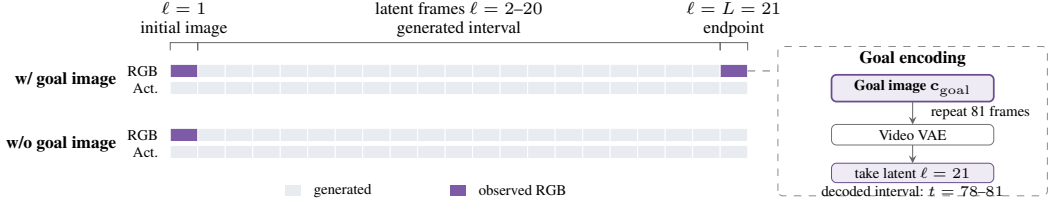
\begin{figure}[!tp]
  \centering
  \resizebox{\textwidth}{!}{%
  \begin{tikzpicture}[x=0.39cm,y=0.55cm,font=\scriptsize]
    \definecolor{generatedgray}{RGB}{232,235,239}
    \definecolor{rgbcondition}{RGB}{125,91,166}

    \draw[black!60] (0,3.95)--(0,4.15)--(1,4.15)--(1,3.95);
    \node[align=center] at (0.5,4.75) {$\ell=1$\\initial image};
    \draw[black!60] (1,3.95)--(1,4.15)--(20,4.15)--(20,3.95);
    \node[align=center] at (10.5,4.75) {latent frames $\ell=2$--$20$\\generated interval};
    \draw[black!60] (20,3.95)--(20,4.15)--(21,4.15)--(21,3.95);
    \node[align=center] at (20.5,4.75) {$\ell=L=21$\\endpoint};

    \foreach \row/\name in {3.20/{w/ goal image},1.55/{w/o goal image}} {
      \node[anchor=east,font=\scriptsize\bfseries] at (-1.65,\row) {\name};
      \node[anchor=east,font=\tiny] at (-0.12,\row+0.22) {RGB};
      \node[anchor=east,font=\tiny] at (-0.12,\row-0.22) {Act.};
      \foreach \slot in {0,...,20} {
        \draw[fill=generatedgray,draw=white,line width=0.25pt]
          (\slot,\row+0.05) rectangle ++(1,0.34);
        \draw[fill=generatedgray,draw=white,line width=0.25pt]
          (\slot,\row-0.39) rectangle ++(1,0.34);
      }
      \draw[fill=rgbcondition,draw=white,line width=0.25pt]
        (0,\row+0.05) rectangle ++(1,0.34);
    }
    \draw[fill=rgbcondition,draw=white,line width=0.25pt]
      (20,3.25) rectangle ++(1,0.34);

    \draw[dashed,black!45,rounded corners=2pt]
      (22.10,0.20) rectangle (31.90,4.05);
    \node[font=\scriptsize\bfseries] at (27.00,3.73)
      {Goal encoding};
    \draw[dashed,black!55] (21.00,3.42)--(22.10,3.42);
    \draw[fill=rgbcondition!12,draw=rgbcondition!85!black,
      rounded corners=2pt,line width=0.7pt]
      (24.05,2.63) rectangle (29.95,3.32);
    \node[font=\tiny\bfseries,align=center] at (27.00,2.98)
      {Goal image $\mathbf{c}_{\mathrm{goal}}$};
    \draw[->,>=stealth,black!60] (27.00,2.63)--(27.00,2.04);
    \node[font=\tiny,anchor=west] at (27.35,2.34) {repeat 81 frames};
    \draw[fill=white,draw=black!55,rounded corners=2pt]
      (24.05,1.51) rectangle (29.95,2.04);
    \node[font=\tiny] at (27.00,1.78) {Video VAE};
    \draw[->,>=stealth,black!60] (27.00,1.51)--(27.00,1.04);
    \draw[fill=rgbcondition!12,draw=rgbcondition!85!black,
      rounded corners=2pt]
      (24.05,0.54) rectangle (29.95,1.04);
    \node[font=\tiny] at (27.00,0.79) {take latent $\ell=21$};
    \node[font=\tiny] at (27.00,0.33) {decoded interval: $t=78$--$81$};
    \draw[fill=generatedgray,draw=none] (4.2,0.20) rectangle ++(0.55,0.28);
    \node[anchor=west,font=\tiny] at (4.90,0.34) {generated};
    \draw[fill=rgbcondition,draw=none] (10.2,0.20) rectangle ++(0.55,0.28);
    \node[anchor=west,font=\tiny] at (10.90,0.34) {observed RGB};
  \end{tikzpicture}
  }
  \caption{\textsc{Plan} conditioning masks and goal encoding. Both settings observe the initial image; the default also fixes a terminal latent encoded from repeated copies of the goal image. Hand and camera trajectories are predicted for all frames.}
  \label{fig:endpoint-conditioning-protocol}
\end{figure}

Both \textsc{Plan} settings receive the task instruction and initial
image. The default additionally conditions on
$\mathbf{c}_{\mathrm{goal}}=\mathbf{I}_T$; the w/o goal image setting uses
$\mathbf{c}_{\mathrm{goal}}=\emptyset$. The goal image fixes the terminal
video latent during sampling, while hand and camera trajectories are
generated jointly. Figure~\ref{fig:endpoint-conditioning-protocol} shows
the conditioning masks.

\paragraph{Encoding the reference images.}
At inference, we repeat the goal image into an 81-frame constant clip
and encode it with the video VAE, starting with a fresh temporal cache.
Its final latent supplies the terminal observation at $\ell=21$.
The initial image is encoded separately in the same way, using its first
latent at $\ell=1$. These encodes use the two reference images alone.
Both observed latents stay fixed during denoising, and the VAE decodes
the entire output video. The terminal latent corresponds to frames
$t=78$--$81$; the intervening frames $t=2$--$77$ are freely generated.
The final column of Figure~\ref{fig:plan-goal-conditioning-qualitative}
shows the supplied goal image as a reference.

During training, observation masks select clean latents from VAE-encoded
training clips. The same mask construction keeps observed positions
clean and excludes them from the video loss. Action supervision uses
the annotations aligned to all video frames, including the terminal
interval. At inference, hand and camera trajectories are predicted from
queries for all frames.

\paragraph{Evaluation interval.}
The VLM evaluates the complete video, including its transition into the
goal segment, with the goal image supplied separately as a reference.
Our qualitative analysis compares object placement before this segment
begins. Appendix~\suppref{app:plan-geometry-consistency} also reports
video--action consistency over the freely generated interval.

\section{Human and VLM Evaluation}
\label{app:human-vlm}

Table~\ref{tab:human-vlm-evaluation-full} reports the complete human and VLM
comparison of \textsc{Plan}-only and unified training, including 95\%
confidence intervals for the gains. Table~\ref{tab:human-vlm-evaluation}
in the main text shows the VLM means.

\begin{table}[!htbp]
  \centering
  \caption{Human and VLM evaluation of \textsc{Plan}-only and unified training. Human scores use 21 pairs across 15 tasks; VLM scores use 188 pairs across 17 tasks. Scores range from 1 to 5. Unified training gains include 95\% bootstrap confidence intervals.}
  \label{tab:human-vlm-evaluation-full}
  \setlength{\tabcolsep}{3pt}
  \begin{papertabular}{lrrrrrr}
    \toprule
    & \multicolumn{3}{c}{Human} & \multicolumn{3}{c}{VLM} \\
    \cmidrule(lr){2-4}\cmidrule(lr){5-7}
    Dimension & \textsc{Plan}-only & Unified training & Gain [95\% CI] & \textsc{Plan}-only & Unified training & Gain [95\% CI] \\
    \midrule
    Action process & 2.23 & \textbf{3.12} & +0.88 [0.37, 1.42] & 1.88 & \textbf{2.89} & +1.01 [0.73, 1.33] \\
    Instruction adherence & 2.18 & \textbf{3.02} & +0.83 [0.43, 1.23] & 2.27 & \textbf{3.22} & +0.94 [0.62, 1.28] \\
    Physical interaction & 1.98 & \textbf{2.90} & +0.92 [0.53, 1.33] & 2.52 & \textbf{3.32} & +0.80 [0.46, 1.13] \\
    Temporal / identity & 2.05 & \textbf{3.30} & +1.25 [0.73, 1.77] & 2.15 & \textbf{3.57} & +1.42 [1.08, 1.76] \\
    Goal transition & 1.72 & \textbf{3.55} & +1.83 [1.33, 2.30] & 1.98 & \textbf{3.29} & +1.30 [1.03, 1.60] \\
    \bottomrule
  \end{papertabular}
\end{table}

\subsection{Human evaluation}
All participants in the human evaluation provided informed consent.
We compare a \textsc{Plan}-only model with a model learned through unified training
after 5,000 steps. Both receive the same initial image, goal image, and task instruction,
and use the same generation settings and seeds. The study covers 17 task
categories, with 188 generated videos per model. Each video contains
81 frames at 16 fps, generated with CFG 2.

Reviewers watch one video at a time alongside its instruction and reference
images, with the method label hidden. After watching the video, they score
five dimensions independently on a 1--5 scale. Videos can be replayed;
insufficient visual evidence is marked unobservable and excluded from the
corresponding score. Method comparisons pair ratings from the same reviewer
account for identical input conditions and generation seeds.

\paragraph{Aggregation and agreement.}
The human means in Table~\ref{tab:human-vlm-evaluation-full} use 21 complete
pairs across 15 task categories. We average seeds within a window, windows
within a source clip, and clips within a task, then weight tasks equally.
VLM means use all 188 generated pairs with the same averaging procedure.
For the unified training gains, we obtain 95\% percentile confidence
intervals from 10,000 bootstrap resamples of task categories, preserving
the paired methods and the within-task averaging above.
Human--VLM agreement is measured by pooled Spearman correlation on
100 videos with both human and VLM ratings. It captures ranking agreement
across both training settings, including their difference in mean quality.
The correlations are $0.528$, $0.371$, $0.515$, $0.597$, and $0.396$
for action process, instruction adherence, physical interaction,
temporal/identity consistency, and goal transition, respectively.

\paragraph{Sensitivity analysis.}
The human-score gains remain positive after removing cases used to develop
the VLM rubric. Excluding the 17 development pairs leaves 86 rated videos
and 17 complete human pairs, with pooled correlations of
$0.491/0.338/0.530/0.631/0.372$ in rubric order. Removing two conditions
with ambiguous instructions or reference images as well gives
$0.498/0.335/0.534/0.639/0.366$ on 85 videos, with the same 17 complete
human pairs.

\subsection{Scoring rubric}
Humans and the VLM use the five dimensions and score anchors in
Table~\ref{tab:human-vlm-rubric}. Scores 2 and 4 indicate intermediate
performance. Action process and instruction adherence are the primary
criteria. Reviewers assess the manipulation sequence and its transition
toward the specified goal, including the required intermediate actions,
visible contact, and consistency of hands and objects.

\begin{table}[!htbp]
\centering
\caption{Shared human/VLM score anchors. Unobservable dimensions are recorded as missing.}
\label{tab:human-vlm-rubric}

\setlength{\tabcolsep}{4pt}
\renewcommand{\arraystretch}{1.15}
\begin{papertabular}{>{\raggedright\arraybackslash}p{.17\linewidth}
                >{\raggedright\arraybackslash}p{.25\linewidth}
                >{\raggedright\arraybackslash}p{.25\linewidth}
                >{\raggedright\arraybackslash}p{.25\linewidth}}
\toprule
Dimension & 1: Severe failure & 3: Partial completion & 5: Good visible performance \\
\midrule
Action process (primary) & No effective action, wrong direction, or arrival through a sudden jump.
& Some necessary actions are visible, but key steps are missing or transitions are flawed.
& Necessary actions occur clearly in a plausible order; intermediate states support the outcome. \\
\addlinespace
Instruction adherence (primary) & Core instruction is not executed, or the stopping condition is seriously violated.
& Some instructions are executed, with visible omissions or constraint violations.
& Observable requirements and constraints are met, with the correct stopping point. \\
\addlinespace
Physical interaction & Severe penetration, self-moving or disappearing objects, or invalid interaction.
& Clear local problems in contact or shape.
& Observed contact, grasping, movement, and shape changes are plausible. \\
\addlinespace
Temporal / identity consistency & Severe jumps or changes in object count or identity.
& Local discontinuities or shape drift.
& Observed hands, objects, and background remain consistent across frames. \\
\addlinespace
Goal transition & No approach to the goal, or sudden replacement at the end.
& Partly natural approach, with clear late corrections or discontinuities.
& Natural transition across the final frames, with the required relationships established. \\
\bottomrule
\end{papertabular}
\end{table}

\subsection{VLM evaluation}
We use \texttt{qwen3.8-max}~\citep{qwen2026qwen38max} at temperature 0 for independent, method-blind
video scoring. Each request contains the initial and goal images, the task
instruction, and time-ordered frames from one anonymous video. We sample
16 uniformly spaced frames and the final eight consecutive frames, yielding
22 frames after deduplication. The denser sampling near the end captures the
transition into the goal state. Frame indices and timestamps accompany the
images.

The judge applies the shared rubric to each dimension and returns a score,
instruction checks, and supporting frame indices. It evaluates the visible
action sequence, marks insufficient evidence as unobservable, and flags
conflicts between instructions and reference images. Returned frame indices
are checked against the supplied frames. Both the human-alignment study
and the external-baseline comparison use independent single-video scores.

\subsection{Comparison with video-generation baselines}
\label{app:plan-vlm-five-method}
Table~\ref{tab:plan-vlm-external-baselines} compares our goal-conditioned
model with Wan2.2 I2V-A14B and Cosmos 3-Nano, each evaluated with and without
a goal image condition. All five settings use the same scoring rubric and
evaluation tasks. The Goal image column indicates whether a goal image is
provided to the model.

Each setting is evaluated on 188 videos across 17 task categories. We use
the frame sampling and judging procedure above. Scores assess progress
through each video, accounting for the native playback rates of 16 fps for
Ours and Wan2.2 and 24 fps for Cosmos 3-Nano. We average seeds, windows, and
source clips in sequence, then give each task equal weight.
Ours-minus-baseline differences are computed on matched inputs and seeds.
We estimate 95\% confidence intervals with 2,000 bootstrap resamples of
source clips within each task, keeping their windows and seeds together.

\begin{table}[!htbp]
  \centering
  \caption{Ours-minus-baseline VLM score differences with 95\% bootstrap intervals. Positive differences indicate higher scores for Ours.}
  \label{tab:plan-vlm-pairwise-ci}

  \setlength{\tabcolsep}{5pt}
  \begin{papertabular}{lccc}
    \toprule
    Baseline & \shortstack{Goal\\image} & Action process & Instruction adherence \\
    \midrule
    Cosmos 3-Nano & \ding{51}
      & $1.311\;[1.190,1.439]$ & $1.036\;[0.874,1.189]$ \\
    Cosmos 3-Nano & \ding{55}
      & $0.865\;[0.661,1.060]$ & $0.922\;[0.657,1.178]$ \\
    Wan2.2 I2V-A14B & \ding{51}
      & $0.944\;[0.743,1.143]$ & $0.892\;[0.669,1.107]$ \\
    Wan2.2 I2V-A14B & \ding{55}
      & $0.471\;[0.245,0.703]$ & $0.620\;[0.364,0.875]$ \\
    \bottomrule
  \end{papertabular}
  \par\vspace{5pt}
  \begin{papertabular}{lcccc}
    \toprule
    Baseline & \shortstack{Goal\\image} & Physical interaction & \shortstack{Temporal / identity\\consistency}
      & Goal transition \\
    \midrule
    Cosmos 3-Nano & \ding{51}
      & $1.161\;[1.047,1.277]$ & $1.799\;[1.686,1.909]$
      & $1.761\;[1.614,1.917]$ \\
    Cosmos 3-Nano & \ding{55}
      & $0.205\;[0.070,0.331]$ & $-0.033\;[-0.162,0.088]$
      & $1.109\;[0.860,1.347]$ \\
    Wan2.2 I2V-A14B & \ding{51}
      & $1.099\;[0.955,1.250]$ & $1.909\;[1.728,2.082]$
      & $1.567\;[1.343,1.783]$ \\
    Wan2.2 I2V-A14B & \ding{55}
      & $0.232\;[0.124,0.337]$ & $0.245\;[0.137,0.354]$
      & $0.663\;[0.438,0.887]$ \\
    \bottomrule
  \end{papertabular}
\end{table}

Table~\ref{tab:plan-vlm-pairwise-ci} shows consistent gains in action
process, instruction adherence, physical interaction, and goal transition
against every baseline. In particular, the action-process and
instruction-adherence intervals are positive for all four comparisons.
Excluding the two conditions with ambiguous instructions or reference
images preserves the four-dimension lead.

\subsection{Real-world execution evaluation}
\label{app:real-world-execution}

In addition to evaluating the generated videos perceptually, we test whether
the predicted manipulation behaviors can be transferred to and executed by
a physical robot. We conduct two real-world experiments in which the
model-generated action sequences are converted into executable robot
trajectories and replayed in the corresponding physical environments.
This evaluation tests whether the generated behavior captures sufficiently
accurate object interactions and action ordering to support downstream
robot execution.

Starting from a generated manipulation video, we reconstruct the relevant
objects and scene and recover their motion over time. The reconstructed
hand motion is then retargeted to the robot hand to obtain an initial
robot trajectory. 
The
resulting action sequence is then deployed on the physical robot without
further modification.

Thus, the complete evaluation pipeline is
\emph{generated video $\rightarrow$ scene and object reconstruction
$\rightarrow$ robot retargeting $\rightarrow$ 
real-world execution}. This follows do-as-I-do \cite{paliwal2026doasido} evaluation paradigm, in which a visual demonstration is
converted into an executable robot behavior.

For both experiments, we report whether execution reproduces the intended
manipulation process and reaches the specified task outcome. These
real-world trials complement the human and VLM evaluations above: whereas
the perceptual metrics measure whether an action appears correct and
physically plausible in the generated video, real-world execution provides
a direct test of whether the predicted interaction can be grounded into a
physically executable manipulation trajectory.

\subsection{Questionnaire interface}
\begin{figure}[!tp]
  \centering
  \includegraphics[width=.96\linewidth]{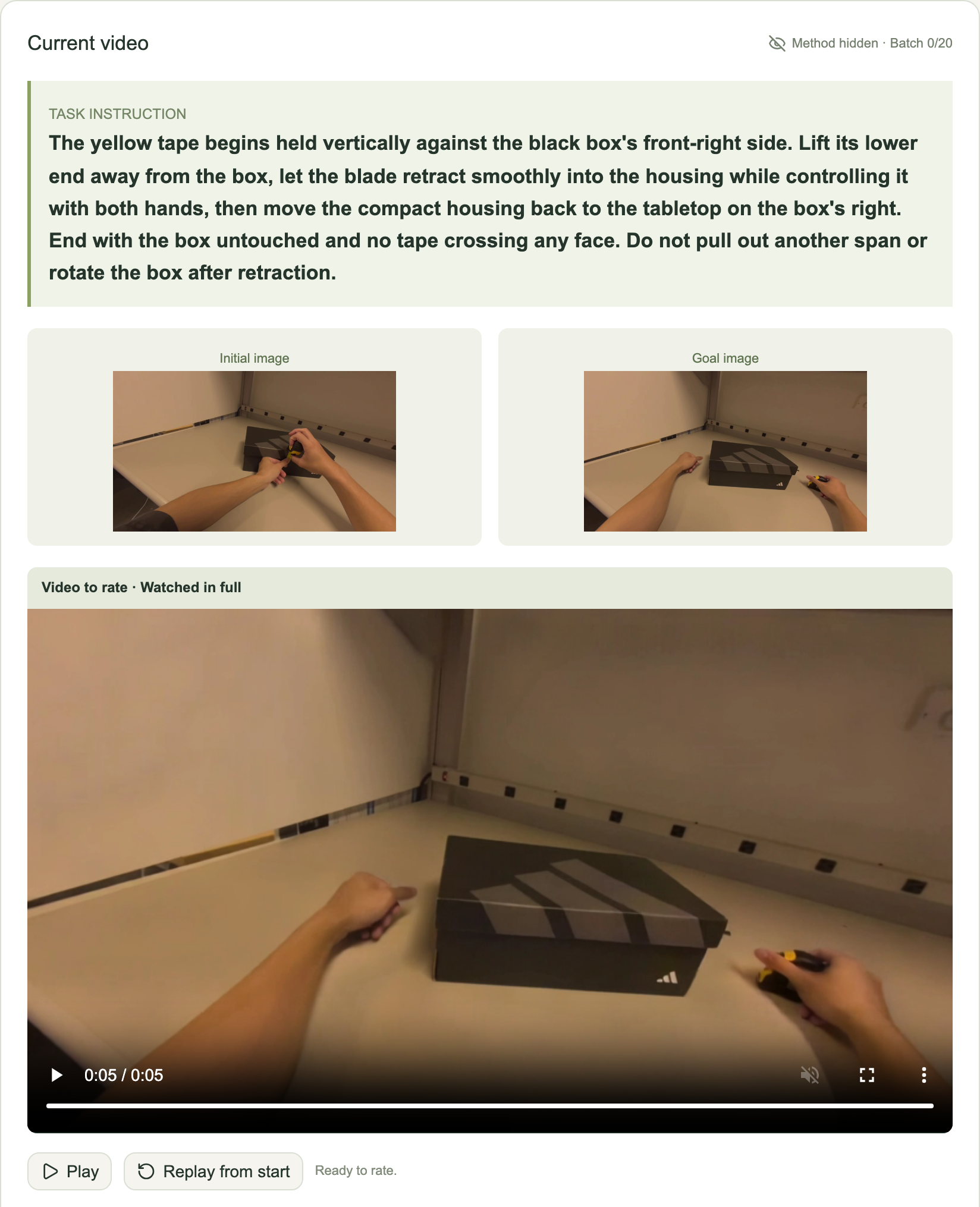}
  \caption{Interface of the questionnaire. Reviewers see the instruction, initial and goal images, and one anonymous video.}
  \label{fig:human-questionnaire-stimulus}
\end{figure}

\begin{figure}[!tp]
  \centering
  \includegraphics[width=.96\linewidth]{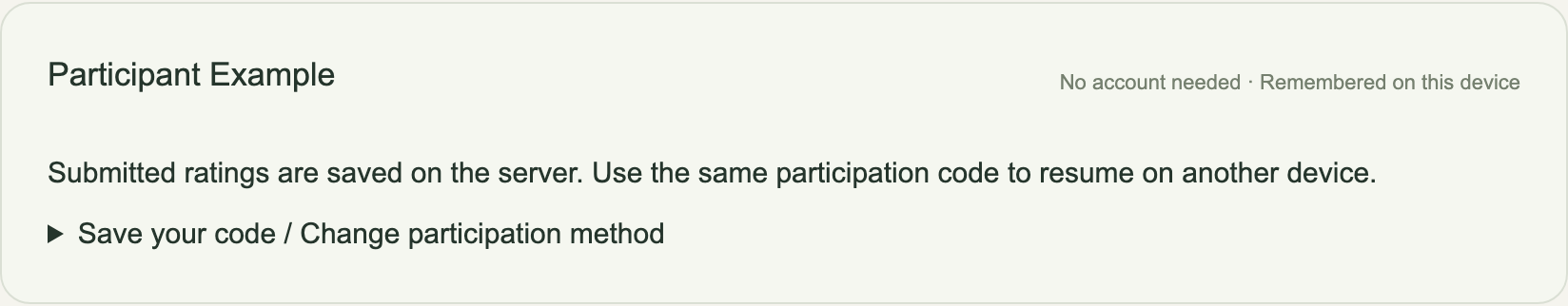}
  \par\vspace{8pt}
  \includegraphics[width=.96\linewidth]{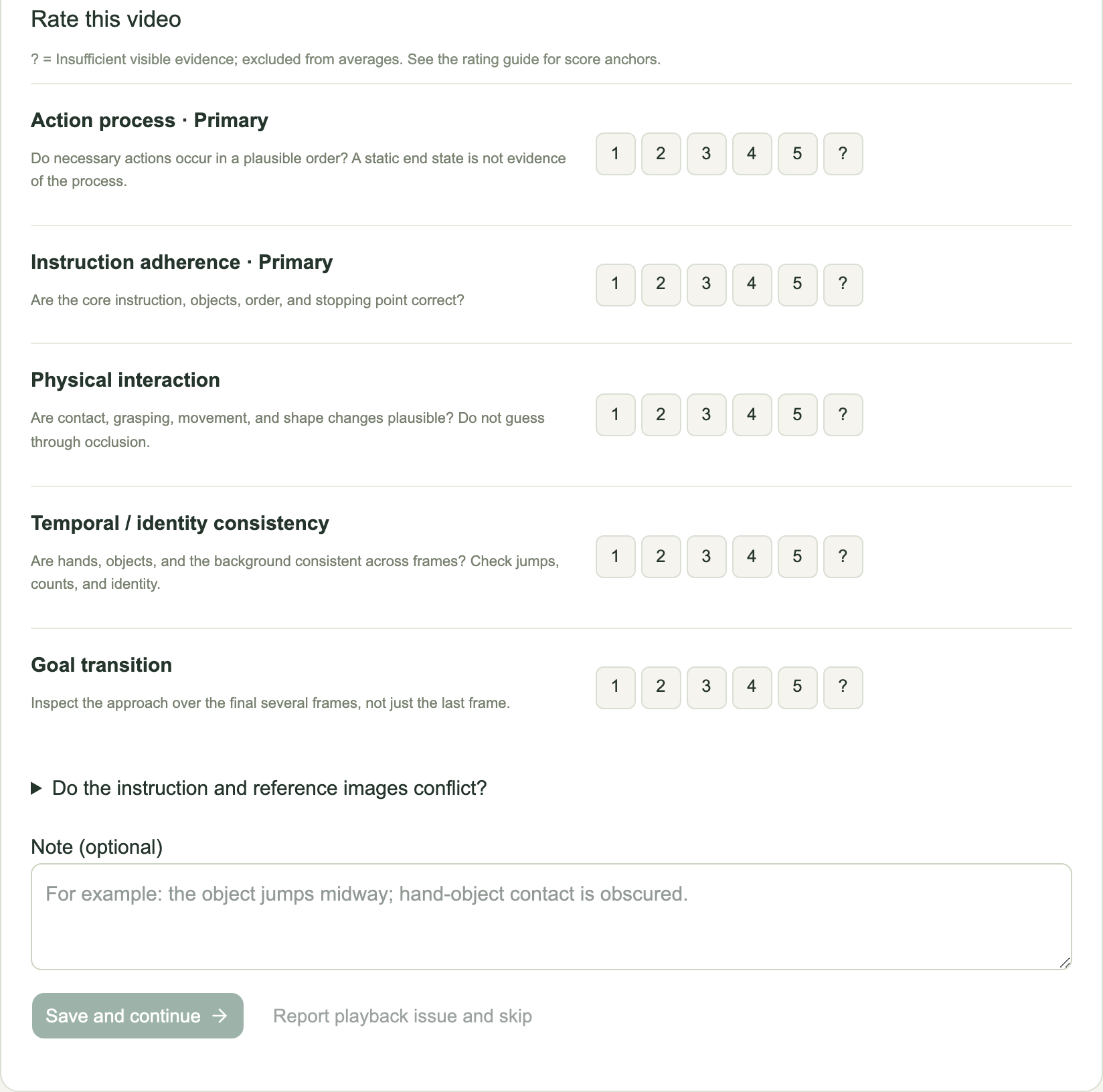}
  \caption{Questionnaire response interface. Each dimension has a 1--5 rating and an unobservable option, with optional notes.}
  \label{fig:human-questionnaire-ratings}
\end{figure}

Figures~\ref{fig:human-questionnaire-stimulus} and
\ref{fig:human-questionnaire-ratings} show the study interface, including
video presentation and the five rating controls.

\section{Additional Quantitative Results}
\label{app:additional_quantitative}

\subsection{\textsc{Perceive} joint errors and hand visibility}
\label{app:perceive-unpenalized}
Table~\ref{tab:reconstruction-unpenalized} reports joint errors from the
same predictions and targets as Table~\ref{tab:reconstruction-external-baselines},
without adding canonical-hand penalties for missed detections.
Our model achieves the lowest mean camera-space error ($47.77$ mm, versus
$95.17$ mm for HaMeR), the lowest root-relative error on every
dataset, and the lowest mean PA error ($9.94$ mm). On HOT3D, HaMeR has lower camera-space error ($71.22$ versus $84.54$ mm); HaMeR also
achieves the lowest PA error on HOT3D and EgoLive.
Our model also achieves the highest visibility F1 on every dataset,
with an overall F1 of $99.17\%$, compared with $92.14\%$ for HaMeR
and $89.95\%$ for Dyn-HaMR.

\begin{table}[!htbp]
  \centering
  \caption{Joint reconstruction and hand visibility in \textsc{Perceive} mode. We report camera-space, root-relative (RR), and Procrustes-aligned (PA) joint errors without missing-hand penalties, alongside separately scored hand-in-frame precision, recall, and F1.}
  \label{tab:reconstruction-unpenalized}

  \setlength{\tabcolsep}{4pt}
  \begin{papertabular}{llrrrrrr}
    \toprule
    & & \multicolumn{3}{c}{Joint error (mm)} & \multicolumn{3}{c}{Hand visibility (\%)} \\
    \cmidrule(lr){3-5}\cmidrule(lr){6-8}
    Dataset & Method & MPJPE $\downarrow$ & RR-MPJPE $\downarrow$ & PA-MPJPE $\downarrow$ & Precision $\uparrow$ & Recall $\uparrow$ & F1 $\uparrow$ \\
    \midrule
    EgoDex-MANO & Ours & \textbf{32.92} & \textbf{31.58} & \textbf{13.12} & \textbf{99.92} & \textbf{100.00} & \textbf{99.96} \\
     & HaMeR & 106.86 & 43.49 & 14.80 & 84.98 & 99.06 & 91.48 \\
     & Dyn-HaMR & 103.58 & 33.82 & 15.05 & 99.52 & 96.86 & 98.17 \\
    \addlinespace[2pt]
    ARCTIC & Ours & \textbf{42.31} & \textbf{13.03} & \textbf{7.57} & 99.93 & \textbf{99.51} & \textbf{99.72} \\
     & HaMeR & 51.36 & 27.36 & 9.42 & 99.23 & 95.47 & 97.31 \\
     & Dyn-HaMR & 202.67 & 21.27 & 9.60 & \textbf{100.00} & 81.62 & 89.88 \\
    \addlinespace[2pt]
    HOT3D & Ours & 84.54 & \textbf{20.11} & 9.33 & \textbf{99.22} & \textbf{97.00} & \textbf{98.10} \\
     & HaMeR & \textbf{71.22} & 45.44 & \textbf{9.32} & 95.21 & 76.40 & 84.78 \\
     & Dyn-HaMR & 92.79 & 28.40 & 11.15 & 95.64 & 71.08 & 81.55 \\
    \addlinespace[2pt]
    EgoLive & Ours & \textbf{31.32} & \textbf{18.70} & 9.76 & 99.55 & \textbf{98.19} & \textbf{98.87} \\
     & HaMeR & 151.25 & 54.90 & \textbf{8.36} & 96.55 & 90.73 & 93.55 \\
     & Dyn-HaMR & 116.80 & 20.46 & 8.63 & \textbf{99.96} & 80.43 & 89.14 \\
    \midrule
    Overall & Ours & \textbf{47.77} & \textbf{20.85} & \textbf{9.94} & \textbf{99.66} & \textbf{98.69} & \textbf{99.17} \\
     & HaMeR & 95.17 & 42.80 & 10.47 & 93.71 & 90.62 & 92.14 \\
     & Dyn-HaMR & 128.96 & 25.99 & 11.11 & 98.96 & 82.45 & 89.95 \\
    \bottomrule
  \end{papertabular}
  \par\vspace{2pt}
  \parbox{0.98\textwidth}{\scriptsize Overall averages joint errors equally across datasets and pools hand-frame decisions for visibility metrics. HaMeR's camera-space translation is fitted to its own 2D projections using dataset intrinsics, with pose and scale fixed; no reference joints enter fitting. Protocols are in Appendix~\suppref{app:perceive-metrics}.}
\end{table}

\subsection{Reconstruction against original EgoDex joints}
\label{app:egodex-original-joints}
We also evaluate all three methods against the original EgoDex 3D joint
annotations, mapped to the common 21-joint hand ordering
(Table~\ref{tab:egodex-original-joints}). We use the same predictions
and evaluation windows as the EgoDex-MANO comparison.
Our model achieves the lowest error in all three coordinate settings,
both with and without presence penalties. Camera-space error is
$30.07$ mm and root-relative error is $19.02$ mm, compared with
$114.58$ and $27.45$ mm for Dyn-HaMR under presence-aware scoring.
PA error is $10.98$ mm, compared with $13.70$ mm for HaMeR.

\begin{table}[!htbp]
  \centering
  \caption{Reconstruction against original EgoDex joint annotations. Errors are in mm (lower is better); $p$ denotes presence penalties.}
  \label{tab:egodex-original-joints}
  \setlength{\tabcolsep}{4pt}
  \begin{papertabular}{lrrrrrr}
    \toprule
    & \multicolumn{3}{c}{Without presence penalties} & \multicolumn{3}{c}{With presence penalties} \\
    \cmidrule(lr){2-4}\cmidrule(lr){5-7}
    Method & MPJPE & RR-MPJPE & PA-MPJPE & MPJPE-\textit{p} & RR-MPJPE-\textit{p} & PA-MPJPE-\textit{p} \\
    \midrule
    HaMeR & 106.23 & 39.75 & 11.98 & 112.30 & 40.90 & 13.70 \\
    Dyn-HaMR & 102.62 & 24.20 & 12.26 & 114.58 & 27.45 & 15.86 \\
    Ours & \textbf{30.07} & \textbf{19.02} & \textbf{10.98} & \textbf{30.07} & \textbf{19.02} & \textbf{10.98} \\
    \bottomrule
  \end{papertabular}
\end{table}

\subsection{\textsc{Plan} video--action consistency}
\label{app:plan-geometry-consistency}
We reconstruct each \textsc{Plan}-generated video with \textsc{Perceive}
and compare the recovered hand joints with the hand trajectory predicted
by \textsc{Plan}. This measures agreement between the two outputs of the
same model. We use four EgoDex image pairs and four generated image pairs,
and average camera-space and wrist-relative joint distances over both
hands, frames, and examples.

The camera-space discrepancy is $4.79$ mm for EgoDex and $4.99$ mm for
generated images; wrist-relative discrepancies are $2.31$ and $2.76$ mm
(Table~\ref{tab:plan-geometry-consistency}). The results remain similar
over frames $2$--$77$, before the goal-conditioned terminal interval.

\begin{table}[!htbp]
  \centering
  \caption{Agreement between \textsc{Plan} hand trajectories and \textsc{Perceive} reconstructions of the generated videos. MPJPE is in mm (lower is better).}
  \label{tab:plan-geometry-consistency}
  \setlength{\tabcolsep}{6pt}
  \begin{papertabular}{lrrrr}
    \toprule
    & \multicolumn{2}{c}{Frames $2$--$81$} & \multicolumn{2}{c}{Frames $2$--$77$} \\
    \cmidrule(lr){2-3}\cmidrule(lr){4-5}
    Image source & MPJPE & RR-MPJPE & MPJPE & RR-MPJPE \\
    \midrule
    EgoDex & 4.79 & 2.31 & 4.78 & 2.32 \\
    Generated images & 4.99 & 2.76 & 4.96 & 2.79 \\
    \bottomrule
  \end{papertabular}
\end{table}

\subsection{Action-guidance sensitivity}
\label{app:action-guidance}
We vary the strength of action conditioning by combining RGB velocity
predictions with and without the prescribed hand and camera trajectories:
\begin{equation}
\hat{\mathbf{v}}_{\gamma}
=\hat{\mathbf{v}}_{\emptyset}
+\gamma\bigl(\hat{\mathbf{v}}_{\mathrm{act}}
-\hat{\mathbf{v}}_{\emptyset}\bigr).
\label{eq:action-guidance}
\end{equation}
Both branches use the same video latent, noise level, text instruction,
and initial image. Here, $\gamma=0$ omits action conditioning,
$\gamma=1$ uses the conditional prediction, and $\gamma>1$ adds
guidance beyond that prediction. The main \textsc{Predict} comparison
in Table~\ref{tab:world-model-external-baselines} uses $\gamma=1$.

Table~\ref{tab:action-guidance-sensitivity} separates the benefit of
action conditioning from additional guidance. Increasing $\gamma$ from
$0$ to $1$ raises full-image PSNR from $20.644$ to $25.327$ dB and hand
PSNR from $16.764$ to $23.724$ dB. Additional guidance at $\gamma=1.5$
raises hand PSNR by $0.175$ dB to $23.899$ dB, while full-image PSNR is
$25.271$ dB. The hand-region improvement occurs on 13 of 20 windows,
with a paired 95\% bootstrap interval of $[+0.002,+0.342]$ dB.

\begin{table}[!htbp]
  \centering
  \caption{Action-guidance sensitivity for \textsc{Predict} on 20 EgoDex windows. Full-image PSNR
  is measured on native $848\!\times\!480$ frames 2--81. Hand PSNR is measured
  inside $128\!\times\!128$ GT-pose crops using GT-only SAM3 hand pixels with
  at least 20\% crop coverage. Higher is better.}
  \label{tab:action-guidance-sensitivity}

  \setlength{\tabcolsep}{7pt}
  \begin{papertabular}{rrr}
    \toprule
    Scale $\gamma$ & Full-image PSNR $\uparrow$ & SAM3 20\% hand PSNR $\uparrow$ \\
    \midrule
    0.0 & 20.644 & 16.764 \\
    0.5 & 23.726 & 21.008 \\
    1.0 & \textbf{25.327} & 23.724 \\
    1.5 & 25.271 & \textbf{23.899} \\
    2.0 & 25.035 & 23.651 \\
    \bottomrule
  \end{papertabular}
\end{table}

\subsection{Backbone comparison with EgoSim}
\label{app:backbone-comparison}
For a fair backbone comparison with EgoSim~\citep{hao2026egosim}, we
conduct an additional experiment by retraining our model from
Wan2.1-Fun-Control with EgoSim's training setting. We evaluate this
model on the benchmarks of Tables~\ref{tab:world-model-external-baselines}
and~\ref{tab:reconstruction-external-baselines}, alongside our main
Wan2.2 model.

\paragraph{Video prediction.}
With the matched Wan2.1 backbone, our model achieves mean PSNR
$18.020$ dB, SSIM $0.6319$, and LPIPS $0.4251$, compared with
$15.384$ dB, $0.5615$, and $0.4974$ for EgoSim
(Table~\ref{tab:backbone-predict}), improving all three
dataset-averaged video metrics. Our main Wan2.2 model reaches
$19.020$ dB, $0.6547$, and $0.3891$, respectively.

\begin{table}[!htbp]
  \centering
  \caption{\textsc{Predict} results with Wan2.2 and Wan2.1 backbones, alongside EgoSim. Full-frame metrics follow Table~\ref{tab:world-model-external-baselines}. Macro averages the four datasets equally. Bold entries denote the best result for each dataset.}
  \label{tab:backbone-predict}
  \setlength{\tabcolsep}{5pt}
  \begin{papertabular}{llrrr}
    \toprule
    Dataset & Method & PSNR $\uparrow$ & SSIM $\uparrow$ & LPIPS $\downarrow$ \\
    \midrule
    EgoDex & Ours (Wan2.2) & \textbf{24.511} & \textbf{0.8462} & \textbf{0.2679} \\
     & Ours (Wan2.1) & 23.540 & 0.8262 & 0.2984 \\
     & EgoSim & 21.331 & 0.8189 & 0.2955 \\
    \addlinespace[2pt]
    ARCTIC & Ours (Wan2.2) & \textbf{14.106} & 0.5785 & \textbf{0.4411} \\
     & Ours (Wan2.1) & 13.921 & \textbf{0.5903} & 0.4437 \\
     & EgoSim & 12.874 & 0.5545 & 0.4968 \\
    \addlinespace[2pt]
    HOT3D & Ours (Wan2.2) & \textbf{15.912} & \textbf{0.4637} & \textbf{0.4990} \\
     & Ours (Wan2.1) & 14.394 & 0.4075 & 0.5672 \\
     & EgoSim & 12.906 & 0.3266 & 0.6389 \\
    \addlinespace[2pt]
    EgoLive & Ours (Wan2.2) & \textbf{21.551} & \textbf{0.7302} & \textbf{0.3483} \\
     & Ours (Wan2.1) & 20.226 & 0.7036 & 0.3912 \\
     & EgoSim & 14.423 & 0.5459 & 0.5582 \\
    \midrule
    Macro & Ours (Wan2.2) & \textbf{19.020} & \textbf{0.6547} & \textbf{0.3891} \\
     & Ours (Wan2.1) & 18.020 & 0.6319 & 0.4251 \\
     & EgoSim & 15.384 & 0.5615 & 0.4974 \\
    \bottomrule
  \end{papertabular}
\end{table}

\paragraph{Hand reconstruction.}
We also report hand reconstruction for this additional run
(Table~\ref{tab:backbone-perceive}). The Wan2.1 model achieves mean
camera-space, root-relative, and PA errors of $51.43$, $18.78$, and
$9.77$ mm, respectively. The corresponding Wan2.2 results are
$48.58$, $20.85$, and $9.96$ mm.

\begin{table}[!htbp]
  \centering
  \caption{\textsc{Perceive} results with Wan2.2 and Wan2.1 backbones. We report ordinary camera-space, root-relative (RR), and Procrustes-aligned (PA) MANO joint errors in millimeters. Wan2.2 entries evaluate the dense predictions from Table~\ref{tab:reconstruction-external-baselines} before visibility filtering and mesh matching. Macro averages the four datasets equally.}
  \label{tab:backbone-perceive}
  \setlength{\tabcolsep}{4pt}
  \begin{papertabular}{llrrr}
    \toprule
    Dataset & Method & MPJPE $\downarrow$ & RR-MPJPE $\downarrow$ & PA-MPJPE $\downarrow$ \\
    \midrule
    EgoDex & Ours (Wan2.2) & 32.92 & 31.58 & 13.12 \\
     & Ours (Wan2.1) & 30.31 & 18.79 & 10.88 \\
    \addlinespace[2pt]
    ARCTIC & Ours (Wan2.2) & 42.47 & 12.99 & 7.55 \\
     & Ours (Wan2.1) & 38.99 & 13.14 & 8.36 \\
    \addlinespace[2pt]
    HOT3D & Ours (Wan2.2) & 87.07 & 20.12 & 9.42 \\
     & Ours (Wan2.1) & 109.99 & 24.86 & 10.05 \\
    \addlinespace[2pt]
    EgoLive & Ours (Wan2.2) & 31.86 & 18.70 & 9.76 \\
     & Ours (Wan2.1) & 26.44 & 18.36 & 9.78 \\
    \midrule
    Macro & Ours (Wan2.2) & 48.58 & 20.85 & 9.96 \\
     & Ours (Wan2.1) & 51.43 & 18.78 & 9.77 \\
    \bottomrule
  \end{papertabular}
  \par\vspace{2pt}
  \parbox{0.98\textwidth}{\scriptsize Wan2.1 root-position errors are $22.63$, $33.48$, $112.07$, and $23.10$ mm on EgoDex, ARCTIC, HOT3D, and EgoLive, respectively; the four-dataset mean is $47.82$ mm.}
\end{table}

\section{Additional Qualitative Results}
\label{app:additional_qualitative}
The qualitative examples cover three complementary settings.
Figures~\ref{fig:wm-qualitative-comparison}
and~\ref{fig:recon-qualitative-comparison} compare generated videos and
reconstructed hands across the four reporting datasets.
Figure~\ref{fig:plan-goal-conditioning-qualitative} compares goal-conditioned
planning, while Figure~\ref{fig:goal-image-control} uses successive visual
targets to specify spatial arrangements in longer tasks.
Figure~\ref{fig:robot-deployment} connects the generated human trajectory
to physical execution. For goal-conditioned examples, motion before the
terminal observed interval shows the generated transition;
Appendix~\suppref{app:plan-goal-protocol} defines this interval.

\subsection{Single-mode versus unified training}
\label{app:single-mode-unified}
The \textsc{Plan}-only baseline is trained with visual subgoals.
At evaluation, both it and the unified model receive the same initial
image, instruction, and goal image, and retain the goal condition throughout
generation. We use the same scoring procedure for both models. The rubric
evaluates the intermediate action sequence and the transition toward the
goal, including contact and object motion.

\begin{figure}[!tp]
  \centering
  \includegraphics[width=\textwidth]{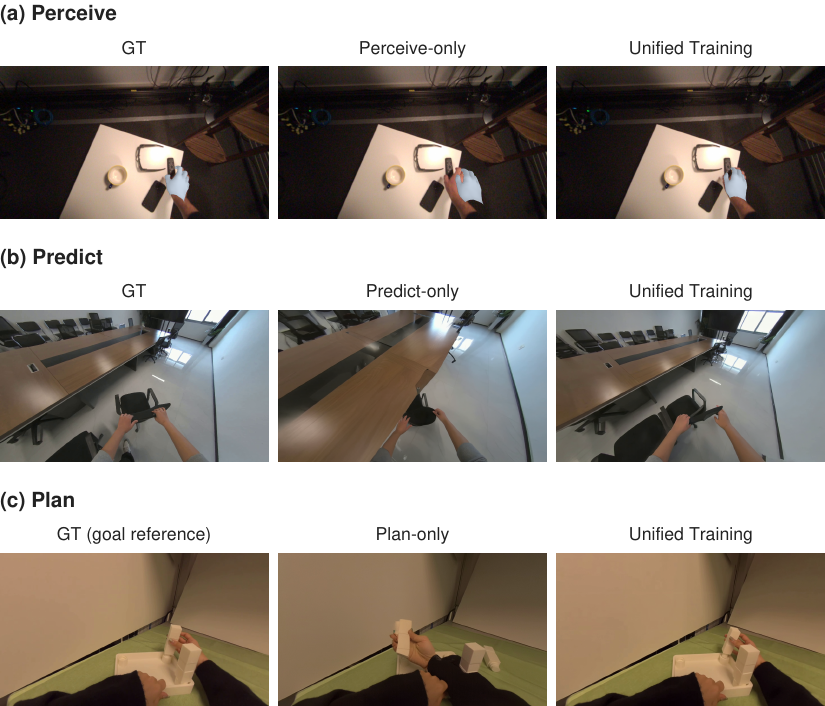}
  \caption{\textbf{Qualitative comparison of single-mode and unified training.}
  \textsc{Perceive}: unified training places the reconstructed palm closer to
  the observed hand.
  \textsc{Predict}: the table extent and viewpoint more closely match GT.
  \textsc{Plan}: the narrow support is positioned beside the fixed broad post.
  The left column shows GT for \textsc{Perceive} and \textsc{Predict}, and
  the goal-image reference for \textsc{Plan}.}
  \label{fig:training-mode-qualitative}
\end{figure}

Figure~\ref{fig:training-mode-qualitative} compares single-mode and unified
training across \textsc{Perceive}, \textsc{Predict}, and \textsc{Plan}.
The selected frames illustrate differences in hand placement, scene
perspective, and object arrangement, respectively.

\FloatBarrier

\subsection{Goal-conditioning ablation and baseline comparisons}
\label{app:plan-goal-ablation}
The default \textsc{Plan} task is conditioned on a visual goal. The
w/o goal image configuration removes this image while retaining the
initial image and instruction; hand and camera trajectories remain outputs.
For multiple windows, visual subgoals are the goal images specifying
intermediate stage targets. Removing the goal image from every window
is distinct from retaining a final visual goal while removing only
intermediate subgoals.

\paragraph{Matched condition-ablation protocol.}
The with/without-goal corpus contains $188$ paired generations over $94$
windows, with two matched seeds per window. Both settings use the same
checkpoint, initial images, instructions, and sampling settings, including
$50$ denoising steps and CFG $2$. Each window starts from its reference
initial image. Only the goal-image condition is removed; the no-goal
model does not receive future hand or camera states.

\begin{figure}[!tp]
  \centering
  \includegraphics[width=\textwidth]{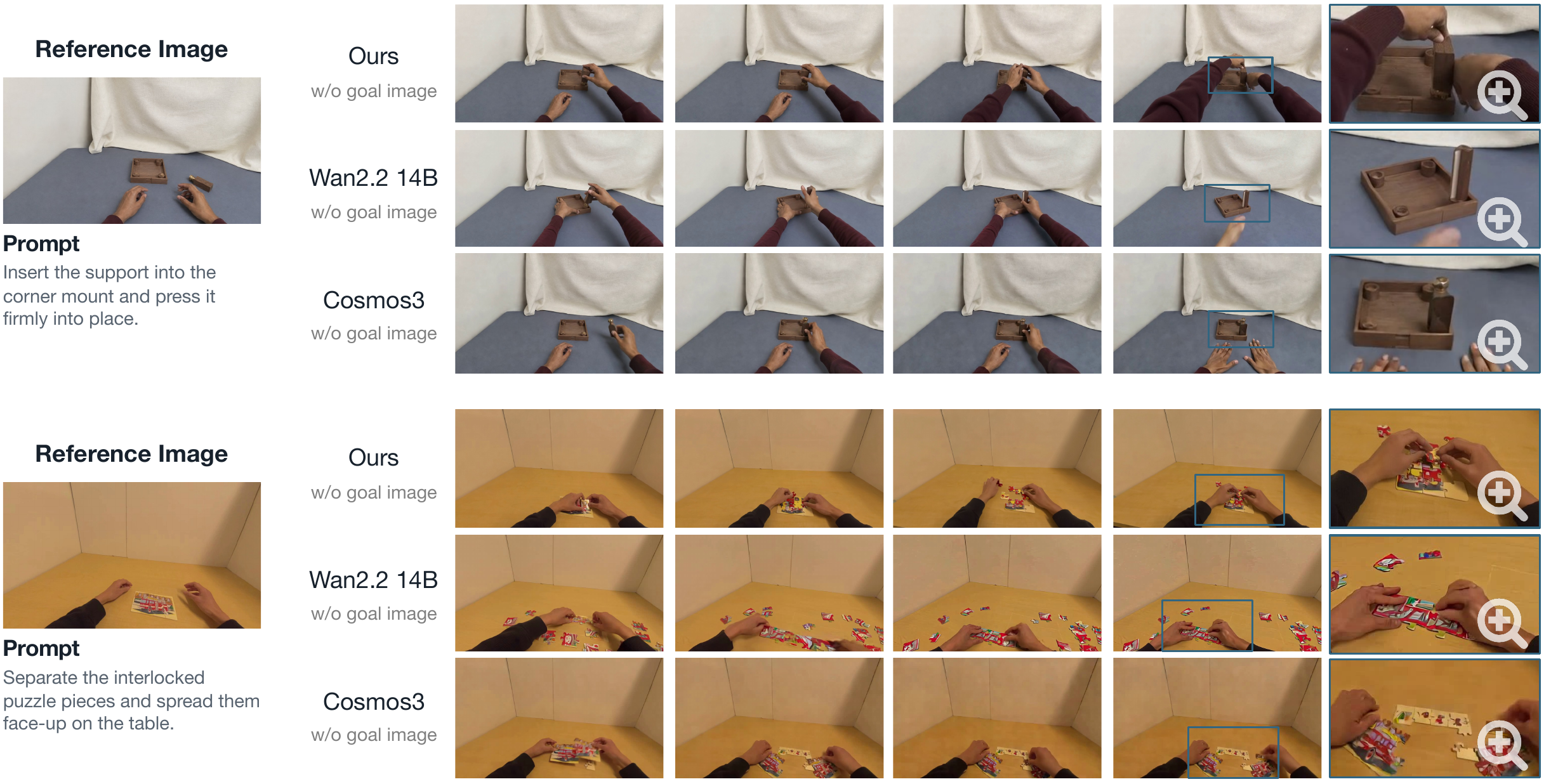}
  \caption{\textsc{Plan} and video-generation baselines without a goal image. All methods receive only the initial image and instruction. Our model maintains the support's shape during placement and the identities of the puzzle pieces during separation. The close-ups show deformations in the baseline outputs.}
  \label{fig:plan-no-goal-qualitative}
\end{figure}

\paragraph{Comparisons without a goal image.}
Figure~\ref{fig:plan-no-goal-qualitative} compares our model with
Wan2.2 and Cosmos when all three methods receive only the initial image
and instruction. Our model maintains the shape of the wooden support
as the hands place it into the corner mount. In the puzzle example,
it keeps individual pieces identifiable as the hands separate them.
Wan2.2 stretches the puzzle pieces, while Cosmos deforms the support
and merges puzzle pieces into elongated strips. These examples show
hand--object consistency during manipulation without a supplied
terminal image.

A visual goal adds control over the desired arrangement beyond the
instruction. Figure~\ref{fig:goal-image-control}, for example, uses the
goal image to specify the separation between the red blocks.

\paragraph{Spatial targets and multiwindow behavior.}
Figure~\ref{fig:goal-image-control} illustrates the task
information supplied by visual goals. The no-goal example places the
blocks differently and the autoregressive rollout accumulates placement
errors. The goal-conditioned windows use prescribed reference starts and
guidance $2$; the no-goal rollout uses generated starts and text guidance
$6$. This execution-strategy comparison is separate from the matched
condition ablation above.
The subgoals provide stage targets; these comparisons do not evaluate
closed-loop error detection or correction.

\FloatBarrier

\section{Responsible Use and Asset Documentation}
\label{app:broader_impact}
\subsection{Source assets and intended use}
The reporting datasets are EgoDex~\citep{hoque2025egodex},
ARCTIC~\citep{fan2023arctic}, HOT3D~\citep{banerjee2025hot3d}, and
EgoLive~\citep{li2026egolive}. Their video, action annotations, and
calibration remain subject to the respective providers' access and
redistribution terms.
Wan model weights and the MANO assets retain their upstream terms.
Derived annotations and visualizations should preserve source attribution;
access-controlled raw data and model assets should be obtained from their
providers.

\paragraph{Intended use.}
\methodname{} supports research on visual prediction, 3D perception, and
robot manipulation. Its benchmark coverage reflects the activities,
cameras, hands, and environments in the source datasets. Performance across
broader populations and embodiments requires additional evaluation.

\subsection{Ethical considerations}
Egocentric video can reveal bystanders, personal spaces, and daily activities.
Use and sharing of these data should follow the source datasets' access and
consent terms, with identifying information minimized in released examples.
Generated interaction videos should be labeled as synthetic. Motion generated
from visual goals requires embodiment-specific validation and supervised
execution before use on physical robots.

\end{document}